\documentclass[preprint,12pt]{elsarticle}

\usepackage{amssymb}
\usepackage{amsmath}
\usepackage{booktabs}
\usepackage{multirow}
\usepackage{arydshln}
\usepackage{xcolor}
\usepackage{tcolorbox} 
\usepackage{float}

\definecolor{darkred}{RGB}{139,0,0}
\definecolor{darkgreen}{RGB}{0,100,0}

\journal{Nuclear Physics B}

\begin{document}

\begin{frontmatter}



\title{KG2Code: Bridging Knowledge Graphs and Large Language Models via Executable Code for Question Answering}

\author[a,b]{Yike Wu\fnref{equal}}
\ead{yike.wu@seu.edu.cn}

\author[a,b]{Nan Hu\fnref{equal}}
\ead{nanhu@seu.edu.cn}

\author[a,b]{Guilin Qi\corref{cor1}}
\cormark[1]
\ead{gqi@seu.edu.cn}

\author[a,b]{Guohui Xiao}
\author[a,b]{Chen Jiang}
\author[a,b]{Xinchun Zou}
\author[a,b]{Yuchen Lu}
\author[a,b]{Songlin Zhai}
\author[a,b]{Yongrui Chen}
\author[c]{Yuyang Zhang}
\author[c]{Xiaoguang Li}
\author[c]{Lifeng Shang}
\author[d]{Jiaoyan Chen}
\author[e]{Jeff Z. Pan}

\fntext[equal]{These authors contribute equally to this work.}

\cortext[cor1]{Corresponding author.}

\affiliation[a]{
    organization={Southeast University},
    country={China}
}

\affiliation[b]{
    organization={Key Laboratory of New Generation Artificial Intelligence Technology and
Its Interdisciplinary Applications (Southeast University), Ministry of Education},
    country={China}
}

\affiliation[c]{
    organization={Huawei Technologies},
    country={China}
}

\affiliation[d]{
    organization={University of Manchester},
    country={United Kingdom}
}

\affiliation[e]{
    organization={University of Edinburgh},
    country={United Kingdom}
}





\begin{abstract}
Recent research has explored the integration of knowledge graphs (KGs) with large language models (LLMs) to enhance their performance on downstream knowledge-intensive tasks, particularly knowledge graph question answering (KGQA). Existing approaches primarily combine LLMs with KGs through retrieval-augmented generation (RAG)-based, agent-based, and SPARQL-based methods. Although these methods have achieved notable success, they still suffer from several limitations, including structural information loss, unfaithful reasoning, and limited flexibility and generalization. To address these challenges, this paper proposes KG2Code, a novel approach that transforms knowledge graphs into a code-based representation, preserving structural semantics while naturally aligning with the code-aware pretraining of modern LLMs. Based on KG2Code, KG2Code-QA is further introduced as a KGQA framework that formulates KGQA as a code generation task. This formulation enables the generation of verifiable reasoning traces and executable code, thereby substantially mitigating the impact of hallucinations. In addition, an automated pipeline is developed to construct a large-scale, high-quality code corpus for effectively training open-source LLMs on KG2Code-QA. After training, LLMs are able to perform KGQA in zero-shot scenarios. Extensive experiments demonstrate that the proposed approach significantly outperforms existing KG-enhanced LLM methods for KGQA, while exhibiting strong generalization to unseen KGs. The code and data are available at Github\footnote{https://anonymous.4open.science/r/test-5074519-5563074/}.
\end{abstract}

\begin{graphicalabstract}
\end{graphicalabstract}

\begin{highlights}
\item KG2Code is the first method to convert KGs into executable code for KGQA reasoning
\item Code-based representation preserves graph structure and fits LLM code abilities
\item We formulate KGQA as code generation with the KG2Code-QA framework
\item KG2Code-QA improves faithfulness, interpretability and generalization
\end{highlights}


\end{frontmatter}



\section{Introduction}
\label{sec1}

The advent of large language models (LLMs) marks a significant milestone in the development of artificial intelligence \cite{DBLP:conf/nips/BrownMRSKDNSSAA20,DBLP:conf/iclr/SanhWRBSACSRDBX22,DBLP:journals/corr/abs-2303-18223}. Owing to their remarkable capabilities in natural language understanding and generation, LLMs have achieved state-of-the-art performance across a wide range of tasks, including question answering, dialogue generation, information extraction, text classification, and machine translation \cite{DBLP:journals/isci/WangYZY25,DBLP:journals/isci/QiuZF26,DBLP:journals/isci/FengLHZCC25,DBLP:journals/isci/TanZJZC26,DBLP:journals/corr/abs-2303-08774}. Nevertheless, they still suffer from issues such as hallucinations, outdated knowledge, and factual inaccuracies \cite{DBLP:journals/csur/JiLFYSXIBMF23,DBLP:conf/acl/MaynezNBM20}, particularly when handling knowledge-intensive downstream tasks such as knowledge graph question answering (KGQA) \cite{DBLP:journals/www/HuWQMCPA23,DBLP:conf/semweb/TanMLLHCQ23}. To address these limitations, recent studies \cite{DBLP:journals/corr/abs-2309-11206,baek-etal-2023-knowledge,sen-etal-2023-knowledge,mavromatis-etal-2025-byokg,DBLP:conf/aaai/TianXLLYCJ025} have explored the integration of external knowledge from knowledge graphs (KGs) into LLMs, with the goal of enhancing their factual consistency and reasoning abilities.

Existing KG-enhanced LLM methods for KGQA can generally be categorized into two paradigms. The first is internal enhancement \cite{DBLP:conf/acl/ZhangHLJSL19,DBLP:journals/tacl/WangGZZLLT21,DBLP:conf/aaai/LiuZ0WJD020}, in which LLMs are trained on corpora constructed from KGs so that KG knowledge is embedded into the model parameters. Under this paradigm, the model is expected to memorize such knowledge and thereby improve performance on downstream question-answering tasks. However, this paradigm has several limitations: it requires training on large-scale corpora, incurs substantial computational costs, and struggles to accommodate frequent knowledge updates \cite{DBLP:journals/corr/abs-2312-10997,baek-etal-2023-knowledge}. To overcome these drawbacks, a second paradigm, namely external enhancement \cite{DBLP:conf/nips/LewisPPPKGKLYR020,DBLP:conf/acl/LuoETPG0MDSLZL24,DBLP:conf/iclr/SunXTW0GNSG24}, has emerged. Instead of requiring LLMs to memorize KG knowledge directly, this paradigm treats KG as an external knowledge source and integrates it with the LLM to enhance question-answering performance.

\begin{figure}[htbp]
\centering
  \includegraphics[width=\linewidth]{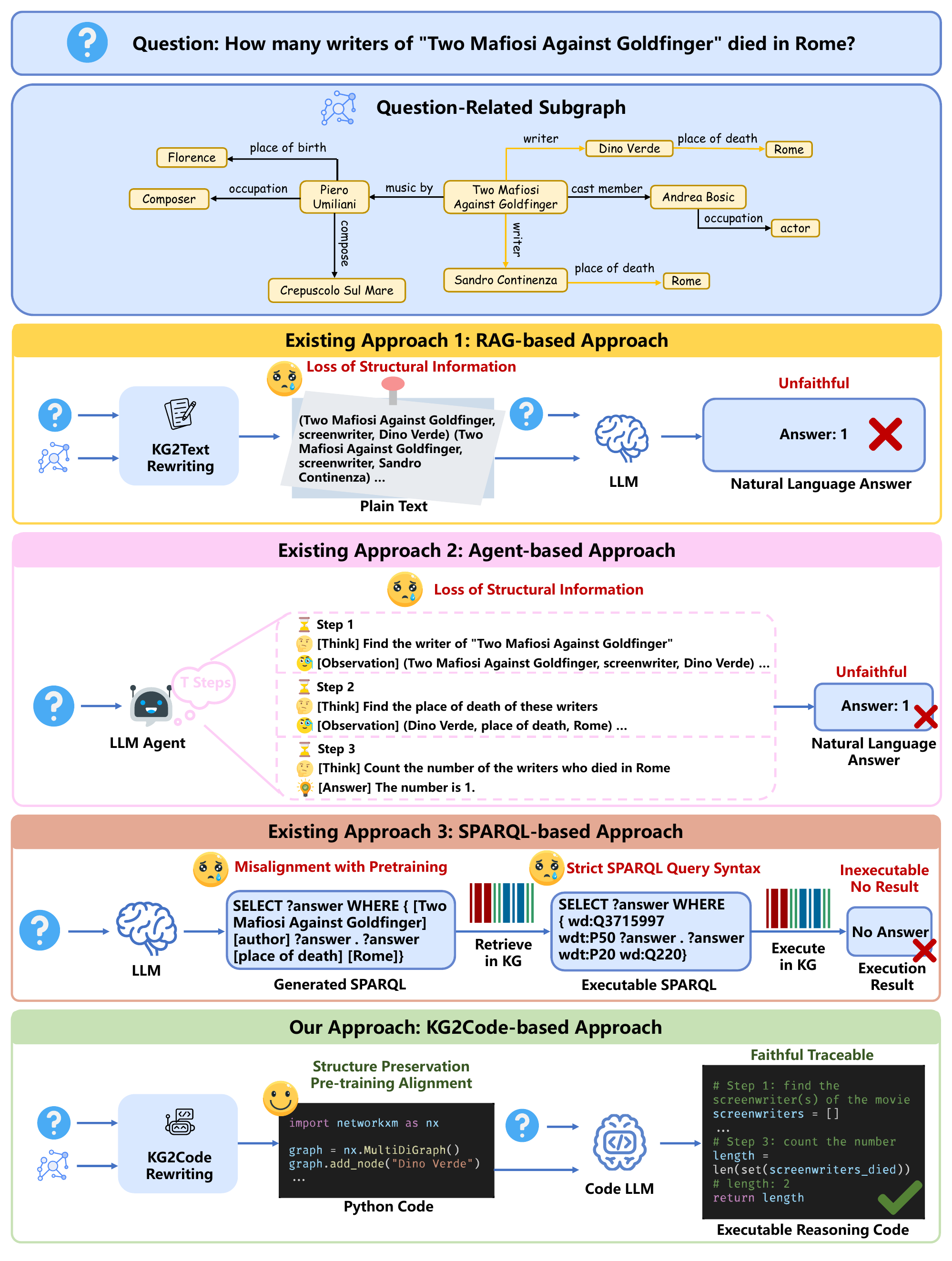}
  \caption{Comparison of existing KG-enhanced LLMs methods for KGQA.}
  \label{fig compare}
\end{figure}

As illustrated in Figure \ref{fig compare}, existing external enhancement methods can be further divided into three categories according to the manner of integration. The first category is retrieval-augmented generation (RAG)-based enhancement \cite{baek-etal-2023-knowledge,DBLP:journals/corr/abs-2309-11206}. These methods retrieve question-relevant triples or subgraphs from the KG, linearize them into textual context, and then concatenate it with the question as input to the LLM. They typically convert graph-structured knowledge into flat text sequences, which inevitably weakens or even loses important structural information, such as entity connectivity, multi-hop dependencies, and relational constraints. Furthermore, even when relevant facts are successfully retrieved, LLMs may still suffer from hallucinations and fail to answer in strict accordance with the retrieved content, thereby producing incorrect responses \cite{DBLP:conf/acl/NiuWZXSZS024,DBLP:conf/iclr/SunZ0XZYSL25}. The second paradigm is agent-based enhancement \cite{DBLP:conf/iclr/SunXTW0GNSG24,DBLP:journals/corr/abs-2404-14741}. Instead of retrieving all relevant knowledge in a single step, these methods allow the LLM to interact with the KG through multiple steps, such as selecting entities, expanding relations, and exploring neighboring nodes. In most cases, the KG is accessed through textual observations or serialized triples, meaning that the complete graph structure is still not directly preserved in the model input. More importantly, the intermediate reasoning process is not guaranteed to be strictly grounded in the KG. Therefore, both RAG-based and agent-based methods suffer from two fundamental limitations: structural information loss and unfaithful reasoning. Although recent studies \cite{DBLP:conf/nips/He0SC0LBH24,DBLP:conf/acl/MavromatisK25} have attempted to incorporate graph neural networks (GNNs) \cite{DBLP:journals/tnn/ScarselliGTHM09} to encode KG structures, such approaches introduce additional graph-specific training and differ from the textual or code-based pretraining paradigm of LLMs, which may limit their generalizability to unseen KGs or tasks. The third paradigm is SPARQL-based enhancement \cite{DBLP:conf/acl/LuoETPG0MDSLZL24,DBLP:conf/coling/FengH25}, which formulates KGQA as a semantic parsing or Text2Query task. This category has limited flexibility and generalization for two main reasons. First, SPARQL has strict syntactic and semantic constraints. Even a minor error can make the query unexecutable or return empty results \cite{DBLP:journals/access/DialloRZ24,DBLP:conf/coling/Mecharniad25}. Second, SPARQL generation is highly dependent on the schema and query style of a specific KG. Models usually require KG-specific training data to generate valid and effective queries.

To overcome the above limitations, this work innovatively leverages code as a bridge between LLMs and KGs. On the one hand, code has become increasingly important in LLMs. It is not only an indispensable component of the training corpus \cite{DBLP:conf/iclr/AryabumiSMMZLFU25,DBLP:journals/jmlr/ChowdheryNDBMRBCSGSSTMRBTSPRDHPBAI23,DBLP:journals/corr/abs-2401-14196}, but also enhances the reasoning capabilities of these models. On the other hand, Python offers a rich ecosystem of graph-related libraries that can faithfully represent graph structures while preserving the inherent topology of KGs. In addition, Python supports a wide range of graph operations, which is particularly well suited for complex KGQA tasks. Building on this insight, we transform KGs into executable code and, based on the code-based representation, formulate KGQA as a code generation task.

Representing KGs as executable code effectively addresses the structural information loss problem by representing KGs in a form that naturally supports graph topology and relational operations. Unlike plain text serialization, code can explicitly encode nodes, edges, attributes, and neighborhood relations, allowing the LLM to perform operations such as traversal, filtering, set comparison, and path-based reasoning over the subgraph. Compared with GNN-based structural encoders, code is better aligned with the pretraining paradigm of LLMs and does not require task-specific graph representation learning. As a widely adopted form of knowledge representation, code has been applied to a variety of KG-related downstream tasks \cite{DBLP:conf/acl/LiZZRLSGLLH0LLY24,DBLP:conf/acl/0001BT0F025,DBLP:journals/talip/BiCJXGCZ24}.

Furthermore, formulating KGQA as a code generation task mitigates the unfaithful reasoning problem by converting KGQA into a code generation and execution process. Because reasoning is carried out through step-by-step operations on graph objects via Python functions, the entire process is inherently grounded in the KG. Consequently, the generated answers remain faithful to the retrieved subgraph. Compared with SPARQL-based methods, code generation for KGQA provides a more flexible and generalizable form of symbolic reasoning. Code constitutes a larger proportion of LLM pretraining corpora than SPARQL \cite{DBLP:journals/corr/abs-2302-13971,DBLP:journals/corr/abs-2308-12950,DBLP:journals/corr/abs-2401-14196}, and has less rigid query-format constraints and is more familiar to modern LLMs due to large-scale code pretraining. The model only needs to generate syntactically valid Python code that operates on the provided graph representation, rather than producing KG-specific SPARQL queries that strictly match a particular ontology or schema.

To this end, this paper proposes KG2Code, a novel KG rewriting method that transforms a KG subgraph into executable Python code. Building on this idea, a KG-enhanced LLM method based on KG2Code is further developed, along with a code-based KGQA framework, namely KG2Code-QA. Our key idea is to use code as an intermediate representation bridging KGs and LLMs. Specifically, the question-relevant KG subgraph is transformed into executable Python code via KG2Code, where entities, relations, and graph connectivity are explicitly represented through data structures and graph operations. The resulting code-style subgraph is then combined with the question and provided to the LLM, which generates executable reasoning code to derive the answer. Inspired by ReAct \cite{DBLP:conf/iclr/YaoZYDSN023}, our framework interleaves natural-language comments and executable Python statements. The comments describe intermediate reasoning intentions, while the code performs concrete operations over the KG-derived data structures. This creative design combines the strengths of existing methods: the natural language comments preserve the reasoning flexibility of LLMs, whereas well-structured code grounds key reasoning steps in executable operations over the given subgraph. In this way, our KG-enhanced method inherits the flexibility and generalizability of RAG-based and agent-based methods, while preserving the graph structure and retaining the faithful KG-grounded reasoning capability of SPARQL-based methods.

Our framework can be applied to both closed-source and open-source LLMs. For closed-source LLMs, KG2Code can be used in a training-free manner, where the model performs code-based KG reasoning with only a few in-context demonstrations. This setting takes advantage of the strong instruction-following and code-generation abilities of advanced closed-source LLMs, while requiring no additional model training. Unlike closed-source LLMs, smaller open-source LLMs often lack reliable code generation abilities for KGQA. They may fail to correctly understand the code-style KG representation, generate invalid graph operations, or produce reasoning programs that cannot faithfully derive answers from the given subgraph. Therefore, high-quality instruction-tuning data are essential for enabling open-source LLMs to perform robust KG2Code reasoning. To address this challenge, an automated corpus construction pipeline is designed to build a large-scale, high-quality KGQA code corpus. The resulting corpus allows open-source LLMs to learn how to understand code-style KGs, conduct graph-grounded reasoning through Python operations, and produce final answers in a verifiable manner.

Extensive experiments are conducted on two widely used KGQA benchmarks, WikiWebQuestions \cite{DBLP:conf/emnlp/XuLCPWSL23} and LC-QuAD 2.0 \cite{DBLP:conf/semweb/DubeyBA019}. The results show that our method consistently outperforms existing KG-enhanced LLM methods for KGQA, including RAG-based, agent-based, and SPARQL-based baselines. These findings demonstrate that representing KGs as code can effectively preserve graph structure, improve faithful reasoning, and better align KG reasoning with the capabilities of LLMs. Furthermore, to evaluate the generalization ability of our framework, transfer experiments are conducted on WebQuestionsSP \cite{DBLP:conf/acl/YihRMCS16} and GrailQA \cite{DBLP:conf/www/GuKVSLY021} under previously unseen KG settings. The strong transfer performance further confirms that KG2Code does not rely on KG-specific query annotations or task-specific graph training, and can generalize effectively across different KGs and QA tasks.

The main contributions of this work are summarized as follows:

\begin{itemize}

\item  To the best of our knowledge, this work makes the first attempt to transform knowledge graphs into executable code for KGQA reasoning. A novel KG rewriting method, KG2Code, is proposed to transform question-relevant KG subgraphs into executable Python code. This code-based representation preserves graph structure, supports flexible graph operations, and provides an LLM-friendly interface for structured KG reasoning.

\item KGQA is formulated as a code-generation and execution task. The proposed framework interleaves natural-language comments with executable Python statements, enabling LLMs to combine flexible reasoning with verifiable, graph-grounded execution. This design mitigates the unfaithful reasoning problem in RAG-based and agent-based methods, while avoiding the rigidity and KG-specific dependence of SPARQL-based methods.

\item An automated corpus construction pipeline is designed for instruction tuning open-source LLMs. The pipeline constructs large-scale, high-quality KGQA code data by converting KG subgraphs into code-style representations, generating reasoning programs, and filtering samples through execution-based verification. As a result, smaller open-source LLMs can acquire reliable code-based KG reasoning abilities.

\item Extensive experiments are conducted on multiple KGQA benchmarks. Results on WikiWebQuestions and LC-QuAD 2.0 show that the proposed method consistently outperforms existing KG-enhanced LLM baselines. Transfer experiments on WebQuestionsSP and GrailQA further demonstrate the strong generalization ability of KG2Code under previously unseen KG settings.

\end{itemize}

\section{Related Work}
\label{subsec1}

\subsection{KG-enhanced LLMs for KGQA}
Early studies on KG-enhanced LLMs mainly focused on integrating KGs during the pre-training or continued pre-training stages of LLMs \cite{DBLP:conf/acl/ZhangHLJSL19,DBLP:journals/tacl/WangGZZLLT21,DBLP:conf/aaai/LiuZ0WJD020}, aiming to improve performance on downstream tasks. However, such approaches are computationally expensive and difficult to scale \cite{DBLP:journals/corr/abs-2312-10997,baek-etal-2023-knowledge}. As a result, recent studies \cite{DBLP:conf/nips/LewisPPPKGKLYR020,DBLP:conf/acl/LuoETPG0MDSLZL24,DBLP:conf/iclr/SunXTW0GNSG24} have increasingly explored the use of KGs as external knowledge sources to enhance the reasoning capabilities of LLMs in downstream applications, particularly in KGQA.

RAG-based enhancement \cite{baek-etal-2023-knowledge,DBLP:journals/corr/abs-2309-11206,DBLP:conf/emnlp/WuHHHQ0P24} adopts the RAG paradigm, where relevant KG facts are retrieved as contextual knowledge to support LLM reasoning on KGQA. KAPING \cite{baek-etal-2023-knowledge} linearizes subgraphs into triples, whereas Retrieve-Rewrite-Answer \cite{DBLP:journals/corr/abs-2309-11206} introduces a knowledge rewriting step to bridge the gap between structured triples and natural language. Subsequent studies \cite{DBLP:conf/emnlp/WuHHHQ0P24,DBLP:conf/emnlp/KoCCYL24} extend this paradigm by incorporating summarization or Chain-of-Thought-guided rewriting, thereby improving the relevance of retrieved knowledge and the structural clarity of the provided context. Although RAG-based methods have shown promising performance across various QA scenarios, they typically linearize graph-structured knowledge into plain textual sequences, which may substantially weaken or even eliminate critical structural information. Structural information plays a crucial role in KGQA, as answering complex questions often requires models to understand and leverage the underlying graph structure. Furthermore, even if evidence pertinent to the question is successfully retrieved, LLMs may still exhibit hallucination, failing to generate responses that faithfully adhere to the retrieved information and consequently producing erroneous answers \cite{DBLP:conf/acl/NiuWZXSZS024,DBLP:conf/iclr/SunZ0XZYSL25}.

Agent-based augmented methods \cite{DBLP:conf/iclr/SunXTW0GNSG24,DBLP:journals/corr/abs-2404-14741,DBLP:conf/aaai/Ma0CSWPTSLZC25} treat LLMs as agents that iteratively explore and reason over KGs. ToG \cite{DBLP:conf/iclr/SunXTW0GNSG24} enables training-free path exploration, while GoG \cite{DBLP:journals/corr/abs-2404-14741} further allows LLMs to generate missing triples during the exploration process. DoG \cite{DBLP:conf/aaai/Ma0CSWPTSLZC25} proposes an iterative KGQA framework that combines subgraph-based answer attempts with multi-role LLM debate, progressively decomposing complex multi-hop questions and improving reasoning reliability over KGs. Compared with standard RAG, these methods provide a more flexible way to access KGs and can better support iterative multi-hop reasoning. Nevertheless, they still rely heavily on the LLM's own planning and generation ability. KGs are typically incorporated into models in the form of textual descriptions or serialized triples, which inevitably ignores the structural information. Moreover, the intermediate reasoning steps generated by the model cannot be reliably ensured to remain strictly grounded in the knowledge graph. The model may choose incorrect exploration paths, ignore retrieved evidence, or generate reasoning steps that are plausible but unsupported by the actual graph. Therefore, the RAG-based method and the agent-based method have two fundamental limitations: structural information loss and unfaithful reasoning. 

SPARQL-based method \cite{DBLP:conf/iclr/YuZNZL0HWWX23,DBLP:conf/acl/LuoETPG0MDSLZL24,DBLP:conf/coling/FengH25} formulates KGQA as a query generation task, where LLMs are prompted or fine-tuned to produce executable queries for retrieving answers from KGs. ChatKBQA \cite{DBLP:conf/acl/LuoETPG0MDSLZL24} first employs a fine-tuned LLM to generate a logical query form, and then applies an unsupervised procedure to replace entities and relations in the query, producing executable SPARQL for answer retrieval. RGR-KBQA \cite{DBLP:conf/coling/FengH25} further improves this pipeline by first retrieving relevant factual knowledge from the KG to enhance the accuracy of logical form generation. It then uses a fine-tuned LLM to generate the logical query and constructs executable SPARQL through entity and relation retrieval. Compared with RAG-based and agent-based methods, SPARQL-based methods offer a clear advantage: they do not need to linearize the KG into plain text for reasoning. Since SPARQL queries are directly executed on the KG, they naturally preserve graph structure and provide faithful, verifiable answers when the generated query is correct. However, this advantage comes at the cost of flexibility and generalization. First, SPARQL queries are governed by rigorous syntactic and semantic rules. As a result, even slight inaccuracies in entity linking, relation identification, variable assignment, or query formulation may cause the generated query to fail during execution or produce no answers \cite{DBLP:journals/access/DialloRZ24,DBLP:conf/coling/Mecharniad25}. Second, the generation of SPARQL queries is strongly influenced by the underlying schema and query patterns of the target KG. Different KGs often use different ontologies, relation names, and structural conventions, so models usually require KG-specific training data to generate valid and effective queries. Such annotated logical forms are expensive to obtain, which weakens the scalability and transferability of SPARQL-based methods. Therefore, while SPARQL-based methods are faithful and structure-preserving, they are less flexible and less generalizable than RAG-based and agent-based methods.

In contrast to the above three lines of research, this work formulates KGQA as a code generation task. Specifically, KG2Code is proposed to transform knowledge graphs into code-style representations, thereby reducing structural information loss and improving the effectiveness of LLM reasoning over graph-structured knowledge. Building on this representation, KG2Code-QA further performs KG reasoning through code generation and execution. By grounding the reasoning process in executable code, this approach alleviates hallucination and faithfulness issues. In addition, compared with SPARQL, code imposes fewer constraints and is more familiar to LLMs. Modeling KGQA as a code generation task also eliminates the need for KG-specific training. To this end, a large-scale, high-quality corpus is constructed to enable trained open-source LLMs to perform KGQA in a zero-shot setting. For more capable closed-source LLMs, reasoning can be conducted directly through in-context learning (ICL) \cite{DBLP:conf/nips/BrownMRSKDNSSAA20,DBLP:conf/emnlp/Dong0DZMLXX0C0S24}, without any additional training.

\subsection{Formulating Downstream Tasks as Code Generation}

Recent LLMs have demonstrated remarkable capabilities in code generation. Building upon this, a series of studies \cite{DBLP:conf/nlpcc/GuoLJLZLLYBGC24,DBLP:conf/acl/LiSTYWHQ23,DBLP:conf/acl/LiZZRLSGLLH0LLY24} have attempted to model downstream tasks as code completion in order to enhance task performance. CODEIE \cite{DBLP:conf/acl/LiSTYWHQ23} reformulates downstream information extraction (IE) as a code completion task, leveraging the structured code generation ability of Code-LLMs to achieve significant improvements over natural language prompting and traditional pretrained IE models in few-shot scenarios. Code4UIE \cite{DBLP:conf/nlpcc/GuoLJLZLLYBGC24} unifies various IE tasks by modeling them as Python class instantiation code completion tasks and integrates an example-retrieval-augmented generation mechanism, providing a general IE framework that is applicable across tasks and datasets. KnowCoder \cite{DBLP:conf/acl/LiZZRLSGLLH0LLY24} models general IE tasks as code completion by uniformly representing heterogeneous schemas as Python classes and adopts a two-phase learning framework, which allows LLMs to understand and follow schemas through code generation, thereby enhancing generalization and extraction capabilities. Beyond IE, CodeTaxo \cite{DBLP:conf/acl/0001BT0F025} addresses taxonomy expansion by casting it as a code completion task, using code-based prompts and similarity filtering to guide LLMs in generating parent-child relationships, effectively improving taxonomy expansion performance in few-shot settings. CodeKGC \cite{DBLP:journals/talip/BiCJXGCZ24} formulates knowledge graph construction as a code completion task by converting natural language into structured code representations and incorporating schema-aware prompting and reasoning-augmented generation, enabling LLMs to better capture semantic structures and enhance knowledge extraction.

In contrast to prior work that generates code from text to construct structured knowledge, KG2Code is introduced to transform knowledge graphs into code and formulate KGQA as a code generation task. This formulation enables executable and verifiable reasoning over retrieved subgraphs, thereby effectively mitigating hallucinations.

\section{Methodology}

\subsection{Preliminaries}

\textbf{Knowledge Graph} (KG) is a structured collection of triples in the form $(s,r,o)$, where $s$, $r$, and $o$ denote the subject, relation, and object, respectively. Formally, it can be represented as $G = \{(s,r,o) \mid s,o \in E,\; r \in R\}$, where $E$ is the set of entities and $R$ is the set of relations.

\textbf{Knowledge Graph Question Answering} (KGQA) aims to answer natural language questions by leveraging a set of facts within KGs. Given a question $q$, the objective is to return an answer set $A(q)\in E$ using the relevant facts in the KG $G$.

\subsection{Overall Framework}

\begin{figure}[htbp]
\centering
  \includegraphics[width=0.75\linewidth]{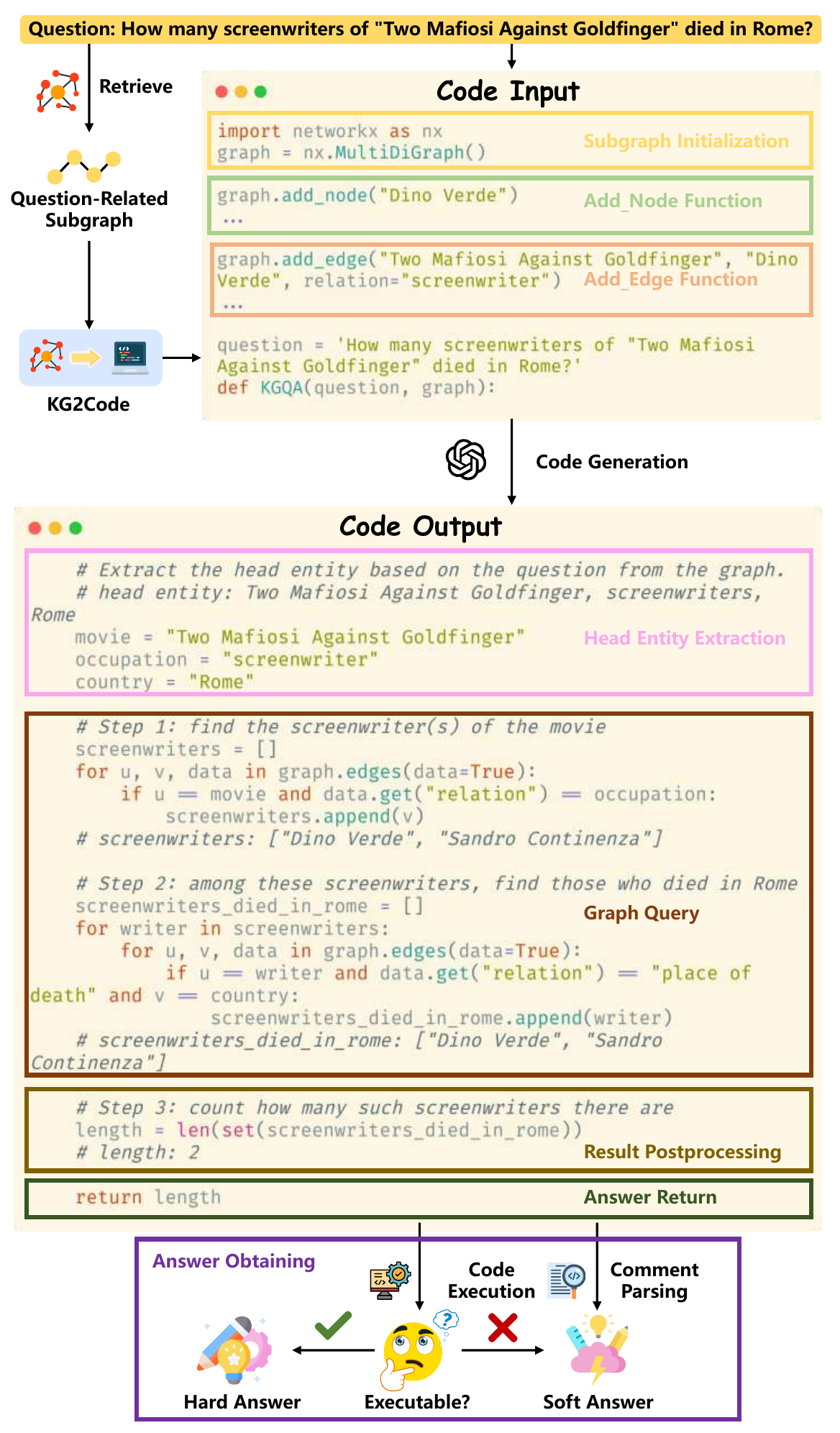}
  \caption{llustration of our framework, KG2Code-QA, highlighting the code-style representation and the code-generation paradigm for the KGQA task. A simplified example is provided for clarity.}
  \label{fig framework}
\end{figure}

The overall architecture of our framework is illustrated in Figure \ref{fig framework}. Given a question, a relevant subgraph is first retrieved from the KG and transformed into executable Python code using KG2Code. The resulting code is then concatenated with the question to form an incomplete code snippet (i.e., denoted as \emph{Code Input} in Figure \ref{fig framework}), which is subsequently fed into the question-answering model. The model completes the code (i.e., denoted as \emph{Code Output} in Figure \ref{fig framework}), and the final answer is obtained by parsing the generated program. 

The two core modules in our framework are KG2Code and Code Generation for KGQA. The role of the KG2Code module is KG rewriting. Specifically, it transforms the retrieved subgraph into executable code using NetworkX. This code constructs a graph object that can be invoked by subsequent code, thereby facilitating a variety of operations on the subgraph. In this way, knowledge rewriting can effectively preserve the structural information of the subgraph.

The Code Generation for KGQA module is designed to complete the KGQA function for answering the question. Its input is an incomplete code snippet formed by concatenating the code representation of the subgraph with the question, and its output is a completed KGQA function. Notably, the KGQA function contains all processing steps required to answer the question; by invoking this function only once, the answer can be directly returned. This function mainly consists of various operations performed on the subgraph in order to answer the question. Since all these operations are implemented using functions provided by Python, the reasoning process can be ensured to be grounded in the KG.

Finally, hard answers are prioritized through code execution, as this approach derives the answer by reasoning over the subgraph and therefore provides stronger guarantees of faithfulness. If the code is not executable, soft answers are extracted from the last line of the comment. Such answers are generated directly by the model and may therefore be affected by hallucinations; nevertheless, this strategy ensures that an answer remains available.

\subsection{KG2Code}

As illustrated in Figure \ref{fig framework}, KG2Code transforms a retrieved knowledge graph subgraph into executable Python code using NetworkX. This representation explicitly preserves the underlying graph structure while naturally aligning with the pretraining corpora of modern LLMs. Formally, each retrieved subgraph is represented as a directed labeled graph $G=(V,E)$, where $V$ denotes the set of entity nodes, $E \subseteq V \times \mathcal{R} \times V$ denotes the set of directed labeled edges, and $\mathcal{R}$ is the set of relation types. Each subgraph is constructed through three core components:

\noindent\textbf{Subgraph Initialization.}
The code first imports the NetworkX library and initializes a task-specific subgraph object (denoted as \texttt{graph} in Figure \ref{fig framework}). This step establishes the structural space in which subsequent node and edge operations are executed, ensuring that the generated code remains consistent with the topology of the retrieved knowledge graph.

\noindent\textbf{Add\_Node Function.}
Given a graph $G=(V,E)$, \texttt{add\_node(v)} inserts a new entity node $v$ into the graph, resulting in an updated graph $G'=(V \cup \{v\}, E)$. Each entity in the knowledge graph is instantiated as a node via \texttt{add\_node()}, with the entity name serving as its unique node identifier. This operation explicitly preserves the existence and identity of each entity in the subgraph. As shown in Figure \ref{fig framework}, the statement \texttt{graph.add\_node("Dino Verde")} adds the entity \texttt{"Dino Verde"} to the graph.

\noindent\textbf{Add\_Edge Function.}
Given a graph $G=(V,E)$, the function \texttt{add\_edge(r, v\textsubscript{head}, v\textsubscript{tail})} adds a directed labeled edge $(v_{head}, r, v_{tail})$ to the graph, resulting in an updated graph $G'=(V, E \cup \{(v_{head}, r, v_{tail})\})$. Each edge is defined by a head entity, a tail entity, and a relation type. This design captures the semantic dependency between entities while preserving the complete topology of the knowledge graph. As shown in Figure \ref{fig framework}, the statement \texttt{graph.add\_edge("Two Mafiosi Against Goldfinger", "Dino Verde", relation="screenwriter")} introduces the relation specified as \texttt{"screenwriter"} between \texttt{"Two Mafiosi Against Goldfinger"} and \texttt{"Dino Verde"}, indicating that \texttt{"Dino Verde"} is the screenwriter of \texttt{"Two Mafiosi Against Goldfinger"}.

Through these components, KG2Code transforms a KG subgraph into structured and executable Python code that can be directly interpreted by LLMs. Unlike textual linearization, this code-based representation preserves structural information and aligns well with LLMs' pretrained code understanding, thereby providing a robust foundation for graph-grounded reasoning in downstream KGQA.

\subsection{KG2Code-QA}


\subsubsection{Code Generation for KGQA}

Beyond KG2Code, a key component of our framework is code-completion-based reasoning. Inspired by ReAct~\cite{DBLP:conf/iclr/YaoZYDSN023}, the model alternates between natural-language comments and executable Python code blocks. The comments explicitly record intermediate reasoning steps, while the code performs concrete operations over the subgraph. This design enforces logical consistency between reasoning and execution, leading to more accurate and interpretable answers. The code completed by the model typically consists of the following components:

\textbullet\; \textbf{Head Entity Extraction.}
Given a question $q$, let $\mathcal{H}(q)=\{h_1,h_2,\dots,$\\e$h_m\}$ denote the set of head entities extracted from $q$, where each $h_i \in V$ corresponds to an entity node in the retrieved knowledge graph. The code first uses comments to explain that the purpose of this block is to identify the head entities in the question. Then, the extracted entities are presented in the form of comments as intermediate results. Finally, each extracted entity $h_i$ is assigned to an individual variable $x_i$, i.e.,
\begin{equation}
x_i \leftarrow h_i, \qquad i=1,2,\dots,m.
\end{equation}
In Figure \ref{fig framework}, the LLM extracts the head entities ``Two Mafiosi Against Goldfinger'', ``screenwriter'', and ``Rome'', and assigns them to the variables \texttt{movie}, \texttt{occupation}, and \texttt{country}, respectively, for subsequent graph queries. This design makes the extracted entities explicit in the code and facilitates their direct reuse in later graph operations.

\textbullet\; \textbf{Graph Query.}
Given a retrieved subgraph $G=(V,E)$, a set of extracted head entities $\mathcal{H}(q)$, and a task-specific query condition derived from $q$, the graph query stage executes a query over $G$ and returns a corresponding result set. Formally, the $t$-th query can be written as
\begin{equation}
\mathcal{Q}^{(t)}(G,\mathcal{H}(q),q)\rightarrow \mathcal{Y}^{(t)},
\end{equation}

where $\mathcal{Y}^{(t)}$ denotes the result of the $t$-th graph query. The code first uses comments to explain the specific objective of the query block, such as retrieving entities that satisfy certain conditions or verifying whether a particular assertion holds. It then iterates over the graph using a \texttt{for} loop to compute
\begin{equation}
\mathcal{Y}^{(t)}=\{\, y \mid y \text{ satisfies the } t\text{-th query condition over } G \,\}.
\end{equation}
Finally, the query results are presented in the form of comments. Such intermediate results are explicitly required, as they not only improve the faithfulness of subsequent code generation but also facilitate the extraction of soft answers. For example, in Figure \ref{fig framework}, the first code block queries the screenwriters of ``Two Mafiosi Against Goldfinger'' and reports the intermediate result \texttt{["Dino Verde", "Sandro Continenza"]} in the comments. The second code block further queries which of these screenwriters died in Rome, and again reports the intermediate result \texttt{["Dino Verde", "Sandro Continenza"]}.

\textbullet\; \textbf{Result Postprocessing.}
Given multiple query results $\mathcal{Y}^{(1)},\mathcal{Y}^{(2)},\dots,$\\$\mathcal{Y}^{(T)}$ produced by the graph query stage, the postprocessing stage applies a task-specific transformation function
\begin{equation}
\mathcal{P}_{\tau}\big(\mathcal{Y}^{(1)},\mathcal{Y}^{(2)},\dots,\mathcal{Y}^{(T)}\big)\rightarrow a,
\end{equation}
where $\tau$ denotes the question type and $a$ represents the final answer. Similar to the graph query stage, code comments are first employed to specify the purpose of the code block. The LLM then identifies the question type and applies the corresponding postprocessing operations to the query results. Finally, the processed results are presented in the form of comments. To accommodate different types of questions, six question types are defined, with the detailed categories listed in Table~\ref{Table type} in Appendix~\ref{appendix type}. The postprocessing operations for each question type are then described in sequence.

\begin{enumerate}
    \item \textbf{Length Calculation.} For counting questions, the final answer is obtained by computing the cardinality of the last query result:
    \begin{equation}
    a = |\mathcal{Y}^{(T)}|,
    \end{equation}
    which is implemented in Python using the \texttt{len()} function.

    \item \textbf{Boolean Evaluation.} For boolean questions, the final answer is obtained by applying logical aggregation over the results of multiple query steps. For example, if each $\mathcal{Y}^{(t)}$ corresponds to a boolean query outcome, the final answer can be written as
    \begin{equation}
    a=\bigvee_{t=1}^{T}\mathcal{Y}^{(t)}
    \qquad \text{or} \qquad
    a=\bigwedge_{t=1}^{T}\mathcal{Y}^{(t)},
    \end{equation}
    corresponding to the Python operators \texttt{or} and \texttt{and}, respectively.

    \item \textbf{Numerical Comparison.} For numerical comparison questions, let $\nu(y)$ denote the numerical attribute associated with a candidate answer $y \in \mathcal{Y}^{(T)}$. The final answer is obtained by filtering the last query result according to a numerical condition:
    \begin{equation}
    a=\{\, y \in \mathcal{Y}^{(T)} \mid \nu(y)\ \theta\ c \,\},
    \end{equation}
    where $\theta \in \{>,<,\ge,\le,=\}$ and $c$ is a task-specific threshold or comparison target.

    \item \textbf{List Sorting.} For extremum-type questions, the last query result is sorted according to an attribute or scoring function $\kappa(\cdot)$:
    \begin{equation}
    \mathcal{Y}'=\operatorname{sort}(\mathcal{Y}^{(T)};\kappa),
    \end{equation}
    and the desired answer is selected from the sorted list, e.g.,
    \begin{equation}
    a=\mathcal{Y}'[0]
    \qquad \text{or} \qquad
    a=\mathcal{Y}'[-1].
    \end{equation}

    \item \textbf{Set Operations.} For set-related questions, the final answer is derived through set operations over multiple query results, such as
    \begin{equation}
    a=\bigcup_{t=1}^{T}\mathcal{Y}^{(t)}
    \qquad \text{or} \qquad
    a=\bigcap_{t=1}^{T}\mathcal{Y}^{(t)}.
    \end{equation}

    \item \textbf{String-level Manipulation.} For string-based questions, let $\sigma(y)$ denote the string form of a candidate answer $y \in \mathcal{Y}^{(T)}$. The final answer is obtained by applying string-level filtering conditions to the last query result:
    \begin{equation}
    a=\{\, y \in \mathcal{Y}^{(T)} \mid \phi(\sigma(y))=\texttt{True} \,\},
    \end{equation}
    where $\phi(\cdot)$ is a string predicate, such as whether the answer begins with the letter ``t''.
\end{enumerate}

In the example shown in Figure \ref{fig framework}, in the final code block, LLMs selects \textbf{Length Calculation} based on the semantics of the question, computes the length of the list as 2, and finally provides the answer in the form of a comment.

\textbullet\; \textbf{Answer Return.} 
The final answer is returned using Python’s \texttt{return} statement. Depending on the question type, the output is a list (factual questions), an integer (counting questions), or a Boolean value (yes/no questions). Detailed examples are provided in \ref{appendix examples}.

\subsubsection{Answer Obtaining}

After code generation, the final answer is obtained using two complementary strategies:

\textbf{Hard answer.} The generated code is executed to obtain the final answer. Since the result is produced through explicit operations on the retrieved subgraph, this strategy provides strong faithfulness guarantees and effectively mitigates hallucinations. In the example shown in Figure \ref{fig framework}, the \emph{Code Input} and \emph{Code Output} are concatenated, followed by appending \texttt{result = KGQA(question, graph)} to invoke the \texttt{KGQA} function. The resulting code is then executed by a Python interpreter, and the value assigned to the variable \texttt{result} is used as the hard answer.

\textbf{Soft answer.} If code execution fails, the answer is extracted by parsing the final natural-language comments in the generated code. This strategy allows the framework to still produce an answer, although with weaker faithfulness guarantees. As shown in Figure \ref{fig framework}, the answer ``2'' is parsed from the last comment line, \texttt{\# length: 2}, and is used as the soft answer.

Each answer type has its own strengths and limitations. Hard answers are more reliable but depend on the executability of the generated code, whereas soft answers are always available but may be affected by hallucinations. Therefore, our framework prioritizes hard answers whenever execution succeeds and falls back to soft answers otherwise. This design balances faithfulness and robustness, ensuring answer completeness for all questions.


\subsection{Code Corpus Generation}
\begin{figure}[h]
\centering
  \includegraphics[width=\linewidth]{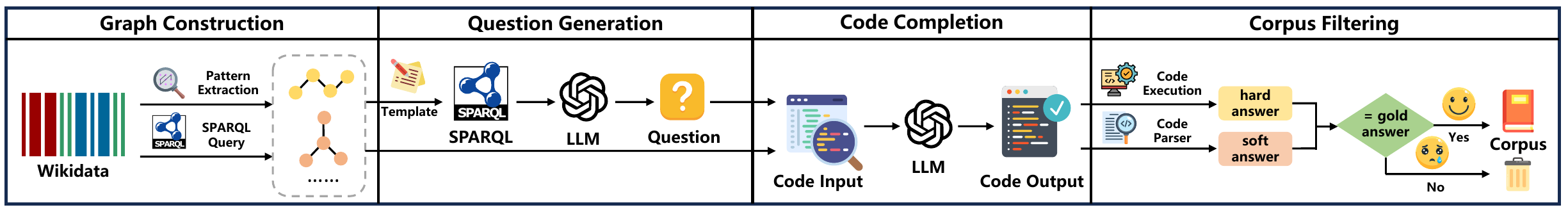}
  \caption{The corpus generation framework.}
  \label{fig corpus}
\end{figure}

To obtain open-source LLMs capable of executing the proposed KGQA method, an automated pipeline is designed for large-scale code corpus construction. As illustrated in Figure~\ref{fig corpus}, the pipeline consists of four stages: subgraph construction, question generation, code completion, and corpus filtering.

\textbullet\; \textbf{Subgraph Construction.}
To ensure the timeliness and accuracy of knowledge, Wikidata \cite{DBLP:journals/cacm/VrandecicK14}, which is continuously updated, is adopted as the underlying knowledge source. Question-relevant subgraphs are constructed through two complementary strategies: pattern extraction and SPARQL querying. In the pattern extraction method, eight types of subgraph structures are predefined, as shown in Table~\ref{Table pattern}, and subgraphs matching these structures are then extracted from Wikidata5M \cite{DBLP:journals/tacl/WangGZZLLT21}. However, direct extraction from Wikidata5M has certain limitations, as it cannot cover all subgraph types involved in KGQA, such as those containing CVT nodes. Therefore, a SPARQL querying method is further adopted. Specifically, a set of entities is first selected as head entities for the questions. Then, different SPARQL queries are designed for different subgraph structures, and the query results are parsed to obtain subgraphs with the corresponding structures. To better simulate real-world application scenarios in which subgraphs may contain irrelevant information, noise is further introduced by randomly performing one-hop triple expansion on entities in the question-relevant subgraphs.

\textbullet\; \textbf{Question Generation.}
Directly generating questions from subgraphs makes it challenging to obtain high-quality questions with broad knowledge coverage. To address this challenge, a two-stage strategy is adopted, in which SPARQL queries are first generated and questions are then derived from these queries. This strategy helps ensure that the generated questions exhibit both sufficient complexity and semantic coherence. Specifically, for subgraphs constructed through pattern extraction, as illustrated in Table~\ref{Table pattern}, the corresponding SPARQL template is selected according to the subgraph type. For subgraphs obtained through SPARQL querying, distinct SPARQL templates are designed for different question types. The selected template is then instantiated with the associated entities and relations to form a SPARQL query. Since the generated SPARQL query is intended for question generation rather than execution, the surface names of entities and relations are directly used instead of their identifiers. Finally, an LLM is employed to generate the corresponding question based on the constructed SPARQL query. The prompt template for question generation is shown in \ref{appendix examples}.

\begin{table*}[h]
\centering
\scalebox{0.6}{
\begin{tabular}{cp{5cm}p{6cm}p{4cm}}
\toprule
\multirow{2}{*}{\textbf{Graph Pattern}} & \multicolumn{3}{c}{\textbf{SPARQL Template}} \\ 
\cline{2-4}
& \multicolumn{1}{c}{\textbf{Factual Question}} & \multicolumn{1}{c}{\textbf{Counting Question}} & \multicolumn{1}{c}{\textbf{Boolean Question}} \\
\midrule
\multirow{2}{*}{\includegraphics[height=0.9cm,keepaspectratio]{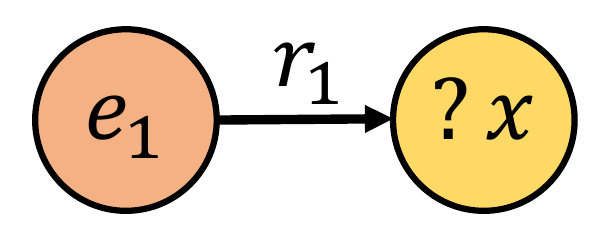}} & SELECT DISTINCT $?x$ \newline WHERE \{ $e_1$ $r_1$ $?x$ . \} & SELECT (COUNT($?x$) as $?count$ \newline WHERE \{ $e_1$ $r_1$ $?x$ . \} &  ASK \{ $e_1$ $r_1$ $?x$ . \}\\
\midrule
\multirow{2}{*}{\includegraphics[height=0.8cm,keepaspectratio]{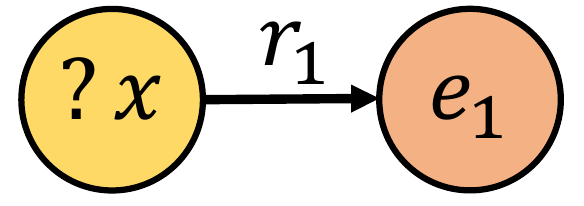}} & SELECT DISTINCT $?x$ \newline WHERE \{ $?x$ $r_1$ $e_1$ . \} & SELECT (COUNT($?x$) as $?count$ \newline WHERE \{ $?x$ $r_1$ $e_1$ . \} & ASK \{ $?x$ $r_1$ $e_1$ . \}\\
\midrule
\multirow{3}{*}{\includegraphics[height=0.8cm,keepaspectratio]{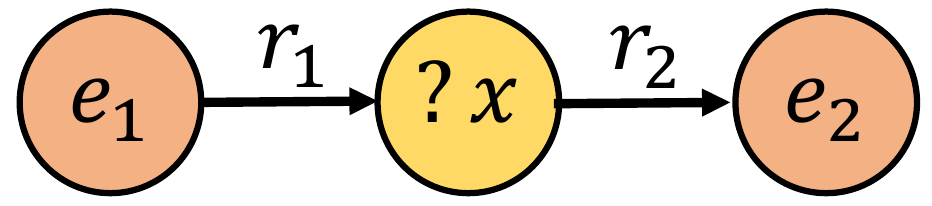}} & SELECT DISTINCT $?x$ \newline WHERE \{ $e_1$ $r_1$ $?x$ . \newline $?x$ $r_2$ $e_2$ . \} & SELECT (COUNT($?x$) as $?count$ \newline WHERE \{ $e_1$ $r_1$ $?x$ . $?x$ $r_2$ $e_2$ . \} &  ASK \{ $e_1$ $r_1$ $?x$ . \newline $?x$ $r_2$ $e_2$ . \}\\
\midrule
\multirow{3}{*}{\includegraphics[height=0.8cm,keepaspectratio]{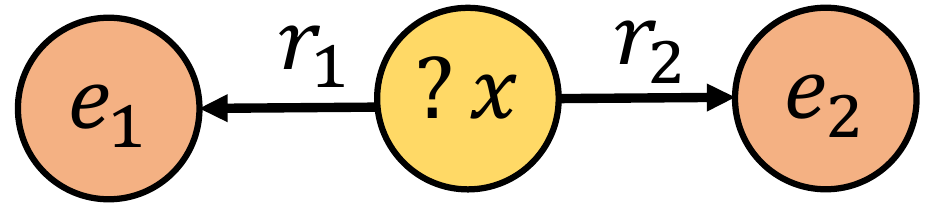}} & SELECT DISTINCT $?x$ \newline WHERE \{ $?x$ $r_1$ $e_1$ . \newline $?x$ $r_2$ $e_2$ . \} & SELECT (COUNT($?x$) as $?count$ \newline WHERE \{ $?x$ $r_1$ $e_1$ . $?x$ $r_2$ $e_2$ . \} &  ASK \{ $?x$ $r_1$ $e_1$ . \newline $?x$ $r_2$ $e_2$ . \}\\
\midrule
\multirow{2}{*}{\includegraphics[height=1.8cm,keepaspectratio]{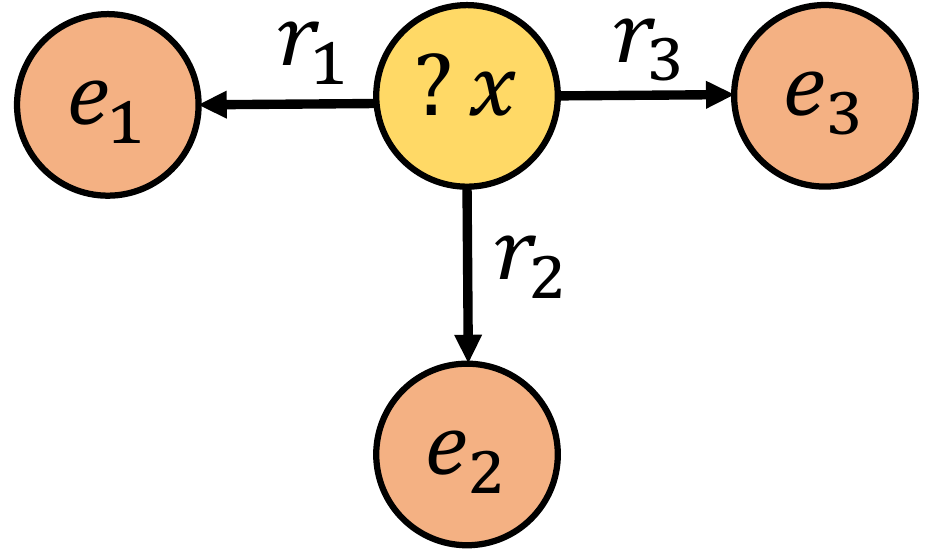}} & SELECT DISTINCT $?x$ \newline WHERE \{ $?x$ $r_1$ $e_1$ . \newline $?x$ $r_2$ $e_2$ . $?x$ $r_3$ $e_3$ . \} & SELECT (COUNT($?x$) as $?count$ \newline WHERE \{ $?x$ $r_1$ $e_1$ . $?x$ $r_2$ $e_2$ . \newline $?x$ $r_3$ $e_3$ . \} & ASK \{ $?x$ $r_1$ $e_1$ . \newline $?x$ $r_2$ $e_2$ . $?x$ $r_3$ $e_3$ . \} \\[1.5cm] 
\midrule
\multirow{3}{*}{\includegraphics[height=0.8cm,keepaspectratio]{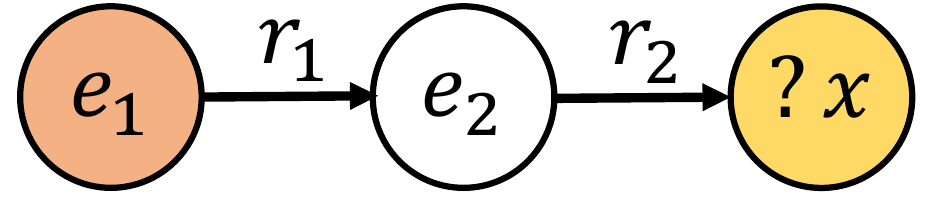}} & SELECT DISTINCT $?x$ \newline WHERE \{ $e_1$ $r_1$ $e_2$ . \newline $e_2$ $r_2$ $?x$ . \} & SELECT (COUNT($?x$) as $?count$ \newline WHERE \{ $e_1$ $r_1$ $e_2$ . $e_2$ $r_2$ $?x$ . \} & ASK \{ $e_1$ $r_1$ $e_2$ . \newline $e_2$ $r_2$ $?x$ . \} \\
\midrule
\multirow{3}{*}{\includegraphics[height=0.8cm,keepaspectratio]{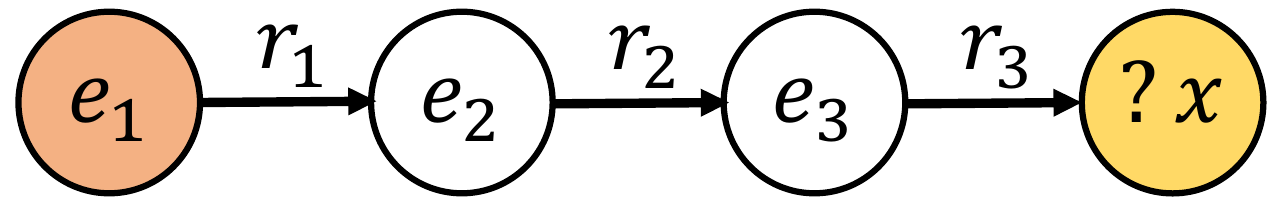}} & SELECT DISTINCT $?x$ \newline WHERE \{ $e_1$ $r_1$ $e_2$ . \newline $e_2$ $r_2$ $e_3$ . $e_3$ $r_3$ $?x$ . \} & SELECT (COUNT($?x$) as $?count$ \newline WHERE \{ $e_1$ $r_1$ $e_2$ . $e_2$ $r_2$ $e_3$ . \newline $e_3$ $r_3$ $?x$ . \} & ASK \{ $e_1$ $r_1$ $e_2$ . \newline $e_2$ $r_2$ $e_3$ . $e_3$ $r_3$ $?x$ . \} \\
\midrule
\multirow{4}{*}{\includegraphics[height=0.8cm,keepaspectratio]{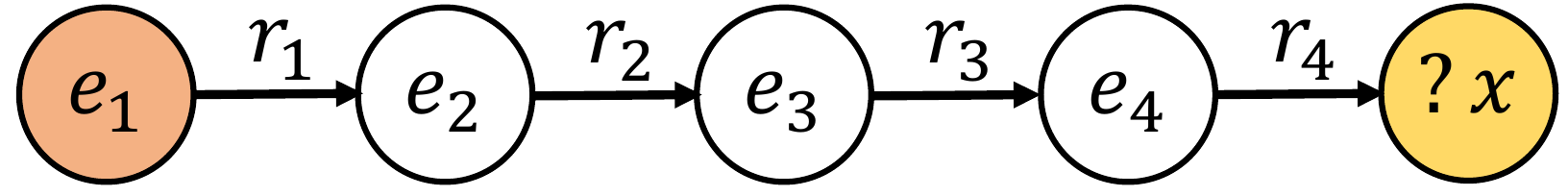}} & SELECT DISTINCT $?x$ \newline WHERE \{ $e_1$ $r_1$ $e_2$ . \newline $e_2$ $r_2$ $e_3$ . $e_3$ $r_3$ $e_4$ . \newline $e_4$ $r_4$ $?x$ . \} & SELECT (COUNT($?x$) as $?count$ \newline WHERE \{ $e_1$ $r_1$ $e_2$ . $e_2$ $r_2$ $e_3$ . \newline $e_3$ $r_3$ $e_4$ . $e_4$ $r_4$ $?x$ . \} & ASK \{ $e_1$ $r_1$ $e_2$ . \newline $e_2$ $r_2$ $e_3$ . $e_3$ $r_3$ $e_4$ . \newline $e_4$ $r_4$ $?x$ . \} \\
\bottomrule
\end{tabular}
}
\caption{Eight question patterns. In the figure, the orange nodes represent head entities, while the golden nodes denote answer entities.}
\label{Table pattern}
\end{table*}

\textbullet\; \textbf{Code Completion.}
For each subgraph-question pair, an incomplete code snippet is constructed by concatenating the KG2Code representation of the subgraph with the corresponding question. This snippet is used as the input, while the completed code is treated as the output. The input-output format follows that of KG2Code-QA introduced earlier. To encourage the generated code to conform to the desired format and reasoning pattern, three in-context demonstration examples are provided, and in-context learning \cite{DBLP:conf/nips/BrownMRSKDNSSAA20} is employed to guide code completion.

\textbullet\; \textbf{Corpus Filtering.} 
The raw code generated by LLMs may contain errors or inconsistencies. To ensure corpus quality, a strict filtering process is applied. For each generated example, the completed code is executed to obtain a hard answer, and the embedded comments are parsed to obtain a soft answer. Only examples for which both answers are correct and consistent with the underlying knowledge graph are retained.

Using this pipeline, over 200,000 high-quality code-based KGQA training examples are collected.

\subsection{Code Corpus Instruction Tuning}

Given the code corpus $D_T={(x_1,y_1),(x_2,y_2),...,(x_T,y_T)}$, the LLM $M_{\theta}$ is trained to generate the target output $y_i$ conditioned on the input $x_i$ by minimizing the negative log-likelihood objective:
\begin{equation}
L = -\frac{1}{T} \sum_{i=1}^T \log p_{\theta}(y_i \mid x_i),    
\end{equation}
where $\theta$ denotes the parameters of $M_{\theta}$ and $p_{\theta}(y_i \mid x_i)$ is the model-assigned probability of producing $y_i$ given $x_i$. 
This training procedure enables the LLM to learn faithful code-based reasoning over knowledge graph subgraphs, serving as the foundation for our KG2Code-QA framework.
 
\section{Experiments}

\subsection{Experiment Setting}

\subsubsection{Datasets}

The evaluation is conducted on two Wikidata-based datasets and two Freebase-based datasets to assess transferability.

\textbf{WikiWebQuestions (WWQ)} \cite{DBLP:conf/emnlp/XuLCPWSL23} is a KGQA benchmark over Wikidata, migrated from WebQuestionsSP \cite{DBLP:conf/acl/YihRMCS16}. It provides manually curated SPARQL annotations with up-to-date answers on Wikidata and includes 2,431 training instances, 454 development instances, and 1,431 test instances. 

\textbf{LC-QuAD 2.0} \cite{DBLP:conf/semweb/DubeyBA019} is a large-scale dataset for complex question answering over knowledge graphs, comprising over 30,000 natural language questions (24,180 training, 6,046 test) paired with executable SPARQL queries over Wikidata and DBpedia 2018. It covers a wide range of complex question types, including multi-hop, temporal, boolean, counting, and qualifier-based queries, and provides paraphrased variants of each question to enhance linguistic diversity and model robustness. Since the official test set does not provide answers, the provided Wikidata SPARQL queries are executed to obtain the answer entities and exclude questions with no returned answers. After this filtering step, the test set contains 4,932 questions.

\textbf{WebQuestionsSP (WebQSP)} \cite{DBLP:conf/acl/YihRMCS16} is a KGQA benchmark that extends WebQuestions with manually annotated semantic parses. Each answerable question is paired with an executable SPARQL query over Freebase, explicitly specifying the topic entity, relations, and constraints required to derive the answers. The commonly used split contains 3,098 training questions and 1,639 test questions. Following \cite{DBLP:journals/corr/abs-2309-11206}, 11 test samples are excluded without answers. 

\textbf{GrailQA} \cite{DBLP:conf/www/GuKVSLY021} is a large-scale KGQA benchmark that consists of 64,331 natural language questions (44,337 train, 6,763 dev, 13,231 test). over Freebase. It includes diverse reasoning types, such as multi-hop reasoning and aggregation operations including counting, comparative, and superlative queries. The training and dev sets provide annotated SPARQL queries and answer entities, while the test set comprises only the questions. For the convenience of evaluation, the development set is utilized for testing.

\subsubsection{Large Language Models}

\textbf{Llama-3.1-8B-Instruct}\footnote{https://huggingface.co/meta-Llama/Llama-3.1-8B-Instruct} is an LLM developed by Meta AI, designed to achieve strong performance across a wide range of natural language understanding and generation tasks. It features improved reasoning capabilities and instruction following compared to previous Llama series models, making it suitable for both research and practical applications.

\textbf{Qwen2.5-Coder-7B-Instruct}\footnote{https://huggingface.co/Qwen/Qwen2.5-Coder-7B-Instruct} is a code-specific variant of the Qwen LLM. Built upon the Qwen2.5 foundation, it has been trained on a large-scale dataset of 5.5 trillion tokens, including source code, text-code grounding data, synthetic data, and other code-related resources. This extensive training enables strong performance in code generation, reasoning, and debugging. In this work, the model is further fine-tuned and adopted as the backbone LLM of KG2Code-QA.

\textbf{GPT-5-Mini}\footnote{https://www.openai.com} is an LLM developed by OpenAI, designed to efficiently perform a wide range of natural language processing tasks. It is a smaller and more optimized version of the larger GPT-5 model, while retaining strong capabilities in generating human-like text, translating languages, summarizing information, and answering questions. In the proposed framework, GPT-5-Mini serves as one of the backbone LLMs of KG2Code-QA. It is also employed to construct the training corpus.

\textbf{GPT-4o-Mini}\footnote{https://www.openai.com} is an advanced language model from OpenAI, built as a compact and efficient variant of the GPT-4 model. It is designed to provide high-level natural language processing capabilities while maintaining lower computational demands. GPT-4o-Mini performs effectively in a variety of tasks, including text generation, question answering, language translation, and content summarization. In this work, GPT-4o-Mini is used to generate the training corpus.

\subsubsection{Baselines}

KG2Code-QA is compared with three categories of KG-enhanced LLM methods for KGQA, as well as CoT-based fine-tuning methods. For SPARQL-based methods, two baselines are established, namely SPARQL-ICL and SPARQL-SFT, to represent the paradigms adopted in recent studies \cite{DBLP:conf/acl/LuoETPG0MDSLZL24,DBLP:conf/coling/FengH25}. For RAG-based methods, direct KG linearization and its variants are employed, which are representative of recent RAG-based KGQA methods. For agent-based methods, ToG, GoG, and DoG are adopted.

\textbf{SPARQL-ICL} uses in-context learning to generate SPARQL queries from natural-language questions. Following the paradigm of prior work \cite{DBLP:conf/iclr/YuZNZL0HWWX23,DBLP:conf/acl/LuoETPG0MDSLZL24,DBLP:conf/coling/FengH25}, a corresponding SPARQL query is first generated for each question. The corresponding entities and relations are then retrieved from the KG based on the generated query, and the entity and relation names in the SPARQL query are replaced with their corresponding IDs, thereby converting it into an executable query. Finally, the executable query is executed over the KG to obtain the answer. To improve the question-to-SPARQL generation performance, the head entity in each question is masked, and three demonstration examples are selected from the training and validation sets according to the semantic similarity between the masked question and candidate examples. This entire process is training-free, enabling a direct comparison with KG2Code-QA.

\textbf{SPARQL-SFT} uses a fine-tuned LLM to generate SPARQL queries from questions. Its overall pipeline is identical to that of \textsf{SPARQL-ICL}; however, at the SPARQL generation stage, supervised fine-tuning data are first constructed from the training and validation sets to fine-tune the LLMs. The fine-tuned model is then used to generate SPARQL queries. This baseline is introduced to further demonstrate that SPARQL imposes relatively strict syntactic constraints and remains insufficiently aligned with LLMs even after fine-tuning.

\textbf{ToG} \cite{DBLP:conf/iclr/SunXTW0GNSG24} treats the LLM as an active agent that iteratively explores and reasons over KG paths via beam search, enabling multi-hop, traceable, and correctable reasoning. Since the open-source LLMs used in the experiments do not possess sufficient reasoning capability to effectively serve as such agents, GPT-5-Mini is adopted as the backbone model. For a fair comparison, \textsf{ToG} is executed on the retrieved subgraphs.

\textbf{GoG} \cite{DBLP:conf/emnlp/XuHC0STL0024} treats the LLM as both a reasoning agent and an implicit knowledge graph, allowing it to iteratively retrieve existing triples and generate missing facts through a Thinking--Searching--Generating process. As the open-source LLMs used in the experiments lack the reasoning ability required for this framework, GPT-5-Mini is also employed as the backbone model. To ensure fairness, \textsf{GoG} is conducted on the retrieved subgraphs.

\textbf{DoG} \cite{DBLP:conf/aaai/Ma0CSWPTSLZC25} proposes an iterative KGQA framework that combines subgraph-based answer attempts with multi-role LLM debate, thereby progressively decomposing complex multi-hop questions and enhancing the reliability of reasoning over KGs. GPT-5-Mini is employed as the LLM backbone, and the framework is evaluated on the retrieved subgraphs.

\textbf{Direct Prompt} represents the most naive KG-RAG method. It takes the retrieved subgraph and the question as joint input to the LLM and directly generates the answer. For this baseline, both open-source LLMs and GPT-5-Mini are employed as backbone models.

\textbf{KG-to-Text} \cite{DBLP:journals/corr/abs-2309-11206} transforms the retrieved subgraph into natural language through a KG-to-Text transformation. The resulting textual description is then used as contextual input for the downstream task.

\textbf{Summary} \cite{DBLP:conf/emnlp/KoCCYL24} produces a concise and task-relevant summary of the retrieved subgraph, which is subsequently provided to the LLM as contextual input.

\textbf{CoTKR} \cite{DBLP:conf/emnlp/WuHHHQ0P24} rewrites knowledge representations through iterative reasoning and summarization, producing more informative and contextually aligned textual knowledge for downstream tasks.

\textbf{G-Retriever} \cite{DBLP:conf/nips/He0SC0LBH24} is implemented based on the official source code on GitHub\footnote{https://github.com/XiaoxinHe/G-Retriever}. This baseline is established to demonstrate that GNN-based methods still suffer from limited generalization ability and remain insufficiently aligned with LLM pretraining.

\textbf{CoT-Tuning} uses the same training corpus as KG2Code-QA, but with a different format. It simply concatenates triples linearly to generate a textual representation of the subgraph, which is then concatenated with the question in a prompt and fed into the LLM to generate the answer. To ensure a fair comparison, the output is required to follow the ReAct format, in which reasoning steps and intermediate results are generated alternately.

The KG2Code-QA framework is also implemented based on GPT-5-Mini. Three examples are used as demonstrations to standardize the output format. This baseline is designed to evaluate whether KG2Code-QA can be effectively applied to closed-source LLMs. Since the KG2Code-QA corpus is also generated by GPT-5-Mini, this baseline further enables an examination of the effects of dataset construction and the fine-tuning of open-source LLMs.

\subsubsection{Training Corpus}
The proposed method is trained exclusively on the constructed training data, rather than on the training split of the benchmark. Therefore, all predictions are made in a zero-shot setting with respect to the benchmark. To ensure a fair comparison, the training data for \textsf{CoT-Tuning} and \textsf{G-Retriever} are also constructed from the training set of \textsf{KG2Code-QA}, using the same training/validation split for both methods. For the \textsf{CoT-Tuning} baseline, the code input is parsed to obtain triples and questions. The triples are linearized into plain text and concatenated with the questions using the prompts in Table \ref{Table answer} as the input. For the output of the corpus, comments are extracted from the code output as responses. In addition, a line prefixed with ``Answer: '' is appended to facilitate answer extraction. For the \textsf{G-Retriever} baseline, the question, answer, head entity, and subgraph are extracted from the code-based training corpus and converted into the format required for model training. For the knowledge rewriting training corpus used in the \textsf{KG-to-Text}, \textsf{Summary}, and \textsf{CoTKR} baselines, 4,000 training instances are constructed following the procedure of CoTKR \cite{DBLP:conf/emnlp/WuHHHQ0P24}, and Llama-3.1-8B-Instruct is subsequently fine-tuned to obtain the corresponding knowledge rewriters. In \textsf{SPARQL-SFT}, because schema differences across knowledge graphs lead to variations in SPARQL queries, a question-to-SPARQL training corpus is directly constructed from the benchmark's training and validation sets, ensuring that training and inference follow the same distribution.


\subsubsection{Retrieval Setting}
To simulate a realistic retrieval scenario, the ground-truth subgraphs are first extracted by parsing the gold SPARQL queries. For each non-answer entity in a ground-truth subgraph, its 1-hop neighborhood is retrieved. Up to 10 triples are then incrementally added until the total subgraph size reaches 100 triples.

\subsubsection{Evaluation Metrics}
Precision, F1, and Exact Match (EM) are reported. Precision measures the proportion of correctly predicted entities among all predicted entities, while Recall measures the proportion of correctly predicted entities among all gold answers. F1 is the harmonic mean of Precision and Recall. Exact Match evaluates whether the predicted entity set exactly matches the gold answer set, regardless of order.

\subsubsection{Implementation Details}
All training is conducted based on LoRA \cite{DBLP:conf/iclr/HuSWALWWC22}. Each model is trained for 10 epochs, and the checkpoint with the best validation performance is selected for evaluation. The batch size, learning rate, LoRA rank, LoRA alpha, and LoRA dropout are set to 128, 1e-4, 64, 128, and 0.05, respectively. The implementation is based on PyTorch\footnote{https://pytorch.org/}, Transformers\footnote{https://huggingface.co/docs/transformers/en/index}, DeepSpeed\footnote{https://huggingface.co/docs/accelerate/en/usage\_guides/deepspeed}, Datasets\footnote{https://huggingface.co/docs/datasets/en/index}, PEFT\footnote{https://huggingface.co/docs/peft/en/index}, and vLLM\footnote{https://docs.vllm.ai/en/latest/}. Experiments are conducted on four A100 GPUs with 40GB memory each and eight H100 GPUs with 80GB memory each.

\subsection{Main Results}

The main results are reported in Table \ref{Table main}. The key findings are summarized as follows. \textbf{(1) KG2Code-QA consistently outperforms all baseline methods across all backbone model settings, demonstrating the effectiveness of the proposed KGQA framework.} The SPARQL-based approach performs poorly. Experimental results show that, even after fine-tuning, a substantial proportion of the generated queries remain either non-executable or return empty results, which constitutes the primary source of errors. This suggests that SPARQL imposes strict syntactic and formatting constraints. \textsf{Direct Prompt} achieves strong performance, demonstrating the effectiveness of the RAG paradigm and indicating that modern LLMs already possess a solid capability to interpret KG triples. The three variants of \textsf{Direct Prompt} (i.e., \textsf{KG-to-Text}, \textsf{Summary}, and \textsf{CoTKR}) provide only marginal improvements and, in some cases, even underperform the vanilla \textsf{Direct Prompt}. This can be attributed to the limited generalization ability of the learned rewriting modules on unseen data, as they may introduce noise during the rewriting process, such as missing or distorted knowledge, thereby degrading downstream question answering performance. Another RAG-based method, \textsf{G-Retriever}, performs substantially worse than \textsf{CoT-Tuning} and \textsf{KG2Code-QA}, although all three methods are trained on the same corpus. This suggests that a considerable gap remains between GNNs and LLMs, highlighting the limitations of GNN-based methods. The agent-based methods \textsf{ToG}, \textsf{GoG}, and \textsf{DoG}, despite relying on closed-source LLMs, exhibit relatively poor performance. This may be due to the complexity of their hand-designed reasoning procedures, which can introduce additional noise into the reasoning process. In contrast, allowing LLMs to reason more directly appears to be more effective. In particular, these agent-based methods often perform unnecessary and repetitive reasoning over the same relations, leading to inefficient exploration and lower answer accuracy compared with simpler and more direct reasoning paradigms. Notably, KG2Code-QA can be directly applied to closed-source LLMs without additional training and achieves a clear performance advantage when using GPT-5-Mini, further validating the generality of the proposed framework. \textbf{(2) CoT-Tuning forms a strong baseline but remains inferior to KG2Code-QA.} Instruction tuning with Chain-of-Thought supervision substantially improves performance over most baselines, indicating that explicit reasoning supervision is beneficial for KGQA. Nevertheless, it consistently underperforms KG2Code-QA, demonstrating the superiority of code-based KG representations over textual linearization and the effectiveness of modeling KGQA as a code generation problem. \textbf{(3) KG2Code-QA with open-source LLMs surpasses its closed-source counterpart, underscoring the quality of the constructed training corpus.} Interestingly, KG2Code-QA trained on the automatically constructed corpus outperforms KG2Code-QA using GPT-5-Mini directly. This result suggests that the corpus generation and filtering pipeline effectively distills high-quality reasoning patterns while eliminating erroneous behaviors from the teacher LLM, leading to stronger generalization in open-source models.

\begin{table}[htbp]
\centering
\small
\scalebox{0.85}{
\begin{tabular}{lccccccc}
    \toprule
     \multirow{2}{*}{\textbf{Methods}} & \multicolumn{3}{c}{\textbf{WWQ}} & &\multicolumn{3}{c}{\textbf{LC-QuAD 2.0}}\\
    \cline{2-4}
    \cline{6-8}
    & \textbf{Prec} & \textbf{F1} & \textbf{EM}& & \textbf{Prec} & \textbf{F1} & \textbf{EM}\\
    \midrule
    \multicolumn{8}{c}{\textbf{\textit{Llama-3.1-8B-Instruct}}}\\
    \textbf{SPARQL-ICL} & 40.38 & 38.03 & 29.98 & &17.17 &16.16 &13.87\\
    \textbf{SPARQL-SFT} & 53.64 & 50.88 & 39.13 & &23.52 &22.12 &18.51\\
    \textbf{Direct Prompt} & 71.47 & 68.90 & 50.10 & & 55.84& 54.18& 42.30\\
    \textbf{KG-to-Text} & 59.83 & 57.28 & 39.20& &41.34& 40.43& 29.08\\
    \textbf{Summary} & 45.67 & 44.85 & 25.79& &39.96 & 39.31& 26.36\\
    \textbf{CoTKR} &58.73 & 54.83&35.29 & & 43.77 & 42.75& 29.68\\
    \textbf{G-Retriever} & 70.62&63.89 &52.97 & &65.17 &59.74 &53.49\\
    \textbf{CoT-Tuning} & \underline{76.50}&\underline{72.42}& \underline{61.01}& &\underline{79.23}& \underline{76.04}& \underline{70.78}\\
    \textbf{KG2Code-QA (Ours)} &\textbf{79.48}&\textbf{78.33}&\textbf{64.08}& &\textbf{81.58}& \textbf{79.26}& \textbf{74.78}\\
    \midrule
    \multicolumn{8}{c}{\textbf{\textit{Qwen2.5-Coder-7B-Instruct}}}\\
    \textbf{SPARQL-ICL} &43.77  &41.10  &32.01  & & 17.69 &16.61  &14.01 \\
    \textbf{SPARQL-SFT} & 54.73 & 52.01 & 39.62 & & 27.93 &  26.76&22.71 \\    
    \textbf{Direct Prompt} &57.04&53.51&42.98& &53.44&50.24& 44.12\\
    \textbf{KG-to-Text} & 59.08&54.74&41.64& &47.80&44.37&36.94\\
    \textbf{Summary} & 64.53&60.59&47.03& &55.50&51.90&44.67\\
    \textbf{CoTKR} &57.56 &52.70 &39.20 &  & 50.66 & 47.27 & 41.02\\
    \textbf{G-Retriever} &67.45 &59.96 &48.92 & &60.74 &55.17 &48.44\\
    \textbf{CoT-Tuning} & \underline{75.75} &\underline{74.11}&\underline{62.89}& &\underline{77.21}&\underline{74.86}&\underline{69.57}\\
    \textbf{KG2Code-QA (Ours)} & \textbf{77.12}& \textbf{76.10}& \textbf{63.24}& &\textbf{81.04}&\textbf{77.97}&\textbf{73.11}\\
    \midrule
    \multicolumn{8}{c}{\textbf{\textit{GPT-5-Mini}}}\\
    \textbf{SPARQL-ICL} & 49.67 & 47.10 & 35.50 & &21.93 &20.59 &17.62\\
    \textbf{ToG} & 40.35 & 43.83 & 10.20 & & 34.88 & 31.99 & 15.39\\
    \textbf{GoG} & 61.13 &61.27 &43.47 & & 57.98&58.78 &51.87\\
    \textbf{DoG} & 67.25 & 66.36 & 51.57 & & 45.26 & 44.00 & 38.39 \\
    \textbf{Direct Prompt}  &\underline{71.49} & \underline{67.87}&\underline{53.18} & & \underline{70.89}&\underline{66.88} &\underline{61.64}\\
    \textbf{KG2Code-QA (Ours)}  &\textbf{75.94} &\textbf{75.06} &\textbf{60.38} & &\textbf{75.79} & \textbf{73.57}&\textbf{69.24}\\    
    \bottomrule
\end{tabular}
}
\caption{The overall results of KG2Code-QA and the baselines. Best results are in bold and second-best are underlined.}
\label{Table main}
\end{table}

\subsection{In-depth Analysis}
This section provides a more in-depth analysis of the main experimental results by comparing \textsf{KG2Code-QA} with various baselines. The results demonstrate that the proposed approach effectively alleviates the limitations of existing KGQA methods.

First, SPARQL-based methods fail to return answers in many cases, either because the generated queries contain errors or because the queries are empty. To examine this issue, \textsf{KG2Code-QA} is compared with two SPARQL-based methods in terms of the proportion of questions for which no answer is returned. As shown in Table~\ref{Table no answer}, \textsf{KG2Code-QA} produces substantially fewer no-answer cases than the SPARQL-based baselines. Although relying solely on hard answers may still lead to empty outputs when the generated code cannot be executed, the two answer-merging strategies introduced in the framework effectively mitigate this issue. Compared with code, SPARQL is more challenging for LLMs to learn and generate accurately. Even after fine-tuning, these models still fail to produce valid answers for a considerable number of questions.

\begin{table}[h]
\centering
\small
\scalebox{0.82}{
\begin{tabular}{lccc}
    \toprule
    \textbf{Methods}& \textbf{WWQ} $\boldsymbol{\downarrow}$ & &\textbf{LC-QuAD 2.0} $\boldsymbol{\downarrow}$ \\
    \midrule
    \textbf{SPARQL-ICL} & 48.57 & &69.85 \\
    \textbf{SPARQL-SFT} & 28.72 & &61.82 \\
    \textbf{KG2Code-QA (Hard)} &10.34 & & 17.58\\
    \textbf{KG2Code-QA (Soft)} & 1.82 & &2.37 \\
    \textbf{KG2Code-QA (Ours)} & \textbf{1.54} & &\textbf{1.68} \\
    \bottomrule
\end{tabular}
}
\caption{No answer rate on WikiWebQuestions and LC-QuAD 2.0. We use Llama-3.1-8B-Instruct as the LLM. Best results are in bold.}
\label{Table no answer}
\end{table}

Moreover, both RAG-based and agent-based methods remain limited in their ability to preserve structural information and support faithful reasoning. To further substantiate this claim, \textsf{Direct Prompt}, \textsf{GoG}, and \textsf{KG2Code-QA} are evaluated under the setting where the retrieved subgraph already contains all ground-truth answers. As reported in Table \ref{Table gd-graph}, \textsf{KG2Code-QA} significantly outperforms both \textsf{Direct Prompt} and \textsf{GoG} on this subset of questions. These results indicate that \textsf{KG2Code-QA} can more effectively comprehend the retrieved subgraphs and accurately extract the correct answers. They further confirm the advantages of the code-based representation and demonstrate that the proposed approach enables more faithful reasoning over the retrieved subgraphs.

\begin{table}[h]
\centering
\small
\scalebox{0.82}{
\begin{tabular}{lccccccc}
    \toprule
     \multirow{2}{*}{\textbf{Methods}} & \multicolumn{3}{c}{\textbf{WWQ}} & &\multicolumn{3}{c}{\textbf{LC-QuAD 2.0}}\\
    \cline{2-4}
    \cline{6-8}
    & \textbf{Prec} & \textbf{F1} & \textbf{EM}& & \textbf{Prec} & \textbf{F1} & \textbf{EM}\\
    \midrule
    \textbf{Direct Prompt} &75.84  &75.61  &59.15 & & 61.96&60.28  &49.59 \\
    \textbf{GoG} &74.30 & 67.46 & 50.95& &71.59 &65.31  &57.23 \\
    \textbf{KG2Code-QA (Ours)} &\textbf{85.14} & \textbf{85.78} &\textbf{75.89} & &\textbf{82.23} & \textbf{80.53} &\textbf{76.92} \\
    \bottomrule
\end{tabular}
}
\caption{The performance on instances where the subgraph contains all ground-truth answers. Llama-3.1-8B-Instruct is used as the LLM for Direct Prompt and KG2Code-QA. GPT-5-Mini is used as the LLM for GoG. Best results are in bold.}
\label{Table gd-graph}
\end{table}

\subsection{Transfer Experiments}

All methods are evaluated on the Freebase-based KGQA benchmarks, WebQSP and GrailQA, to assess cross-KG generalization. The results are reported in Table \ref{Table transfer}. The main observations are as follows.

\textbf{(1) KG2Code-QA achieves the best overall performance on both datasets, demonstrating strong generalization to unseen KGs.} Although \textsf{CoT-Tuning} is trained on the same QA corpus, it consistently underperforms KG2Code-QA, highlighting the advantage of code-based reasoning when transferring to new KG schemas and data distributions. Notably, the performance of SPARQL degrades substantially compared with its results on the two Wikidata-based datasets. This decline can be primarily attributed to the greater complexity of SPARQL queries in the Freebase setting, which leads to a lower execution success rate for the generated queries. \textbf{(2) KG2Code-QA yields the largest improvements in Exact Match (EM), indicating more faithful and complete answer generation.} As the strictest evaluation metric, EM measures whether the predicted answer set exactly matches the gold-standard answers. The substantial gains in EM suggest that KG2Code-QA facilitates precise, graph-grounded reasoning, thereby effectively reducing hallucinations and producing more accurate answers in transfer settings.

\begin{table}[h]
\centering
\small
\scalebox{0.85}{
\begin{tabular}{lccccccc}
    \toprule
     \multirow{2}{*}{\textbf{Methods}} & \multicolumn{3}{c}{\textbf{WebQSP}} & &\multicolumn{3}{c}{\textbf{GrailQA}}\\
    \cline{2-4}
    \cline{6-8}
    & \textbf{Prec} & \textbf{F1} & \textbf{EM}& & \textbf{Prec} & \textbf{F1} & \textbf{EM}\\
    \midrule
    \textbf{SPARQL-ICL} & 18.13 &  17.85&16.65  & &15.92 &15.57 &15.16\\
    \textbf{SPARQL-SFT} & 18.18 &17.84  &16.34  & &22.54 & 22.52&21.68\\
    \textbf{ToG} & 38.69 & 43.67 & 10.32 & & 46.53 & 49.76 & 21.38\\
    \textbf{GoG} &69.72 &70.37 & 54.91& &64.51 & 64.99&58.79\\ 
    \textbf{DoG} & 82.33 & 80.76 &\underline{73.65} & &57.48 & 55.89 & 50.06\\
    \textbf{Direct Prompt} & 79.56 & 77.97 & 63.39 & & 69.32& 68.26& 55.06\\
    \textbf{KG-to-Text} & 59.42 & 57.79 & 40.97& &50.22& 48.82& 38.31\\
    \textbf{Summary} & 59.22 & 56.96 & 41.28& &46.88 & 45.71&32.84\\
    \textbf{CoTKR} &66.05 & 62.63&46.13 & &53.01 & 51.77& 38.49\\
    \textbf{G-Retriever} &76.01 &70.11 &59.77 & &75.58 &69.29 &61.85\\
    \textbf{CoT-Tuning} & \underline{85.68}&\underline{82.89}& 73.53& &\underline{77.95}& \underline{75.79}& \underline{71.94}\\
    \textbf{KG2Code-QA (Ours)} &\textbf{85.98}&\textbf{84.78}&\textbf{77.15}& &\textbf{79.84}& \textbf{78.14}& \textbf{73.44}\\   
    \bottomrule
\end{tabular}
}
\caption{Transfer experiment results. ToG, GoG, and DoG are evaluated with GPT-5-Mini, while all other methods use LLaMA-3.1-8B. Best results are in bold and second-best are underlined.}
\label{Table transfer}
\end{table}

\subsection{Hard Answer vs Soft Answer}
Our framework produces two types of answers: hard answers, obtained by executing the generated code, and soft answers, extracted by parsing the natural-language comments embedded in the code. Hard answers are executable, verifiable, and fully grounded in KG operations, while soft answers offer flexibility when code execution is not feasible.

\subsubsection{Answer Executability Analysis}
\begin{figure}[h]
\centering
  \includegraphics[width=\linewidth]{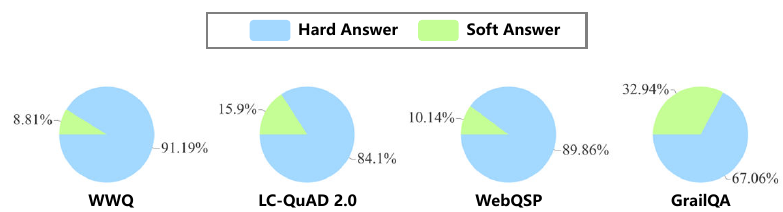}
  \caption{Proportion of hard and soft answers produced by KG2Code-QA across four datasets.}
  \label{fig hard-soft}
\end{figure}

To assess the executability and verifiability of KG2Code-QA, the proportions of hard and soft answers produced by the framework are examined. Using LLaMA-3.1-8B-Instruct as the backbone model, the distribution of answer types across four datasets is analyzed, as shown in Figure \ref{fig hard-soft}. KG2Code-QA predominantly generates hard answers, indicating its ability to reliably generate executable code and perform faithful reasoning over knowledge graph subgraphs. This finding suggests that most predictions are verifiable and largely free from hallucination. In the few cases where code execution fails, the framework effectively falls back to soft answers extracted from comments, thereby ensuring robustness and answer completeness.

\subsubsection{Performance of Hard Answer and Soft Answer}
The performance of different answer types selected as the final output is further compared. Table \ref{Table answer} reports the results under the setting where LLaMA-3.1-8B-Instruct is used as the backbone model. The following observations can be made. \textbf{(1) The proposed strategy that combines hard and soft answers achieves the best overall performance}, outperforming strategies that rely solely on either hard or soft answers. This demonstrates the effectiveness of the fallback mechanism, which leverages execution-based verification when possible while maintaining robustness through comment-based answers. \textbf{(2) Using soft answers alone consistently outperforms using hard answers alone}, suggesting that LLMs may fail to generate fully executable code even when the underlying reasoning is correct. In addition, hard answers are observed to outperform soft answers on questions involving multiple answer entities. In such cases, soft answers may omit some entities, whereas hard answers derived from explicit graph operations are more likely to return complete answer sets. This further highlights the advantage of executable, graph-grounded reasoning over purely text-based reasoning, which may not fully cover the answer space.

\begin{table}[h]
\centering
\small
\scalebox{0.82}{
\begin{tabular}{lccccccc}
    \toprule
     \multirow{2}{*}{\textbf{Methods}} & \multicolumn{3}{c}{\textbf{WWQ}} & &\multicolumn{3}{c}{\textbf{LC-QuAD 2.0}}\\
    \cline{2-4}
    \cline{6-8}
    & \textbf{Prec} & \textbf{F1} & \textbf{EM}& & \textbf{Prec} & \textbf{F1} & \textbf{EM}\\
    \midrule
    \textbf{KG2Code-QA (Hard)} &74.42& 73.82 & 60.52& &70.05& 68.58 & 65.11\\
    \textbf{KG2Code-QA (Soft)} &{78.75}&{75.34}&60.80 & &{79.42} & {76.13} & 70.56\\
    \textbf{KG2Code-QA (Ours)} &\textbf{79.48}&\textbf{78.33}&\textbf{64.08}& &\textbf{81.58}& \textbf{79.26}& \textbf{74.78}\\
    \bottomrule
\end{tabular}
}
\caption{Performance comparison of hard answers, soft answers, and the hybrid answer-obtaining strategy.}
\label{Table answer}
\end{table}

\begin{figure}[h]
\centering
  \includegraphics[width=0.9\linewidth]{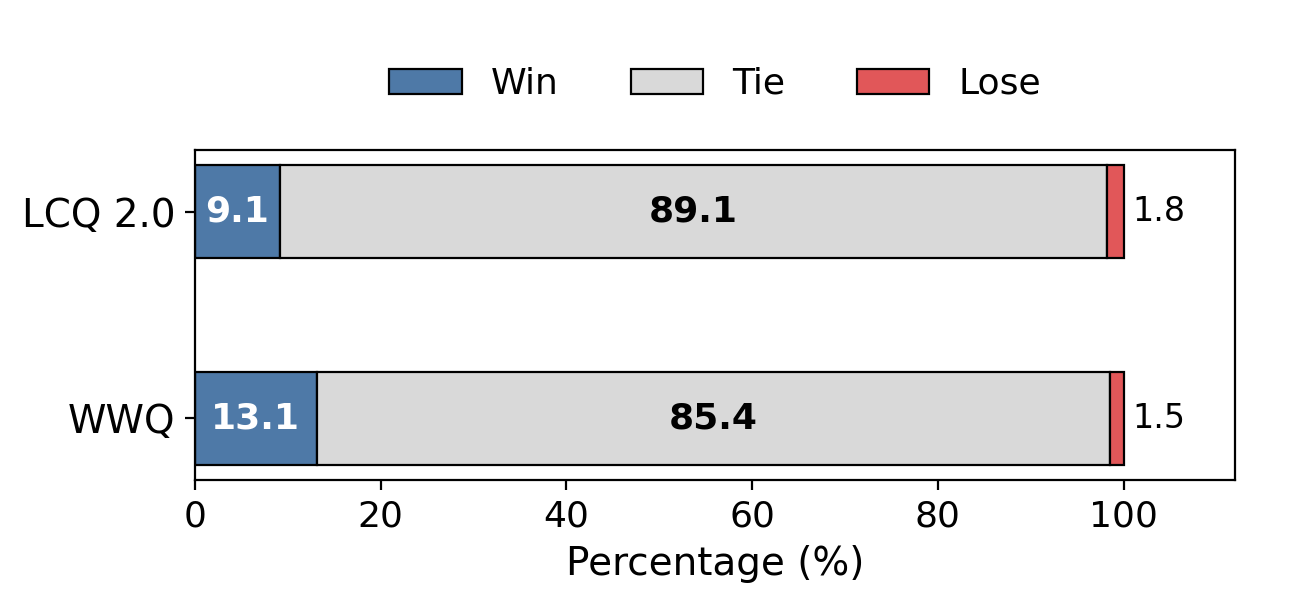}
  \caption{Win–Tie–Lose diagrams comparing the use of hard answers versus soft answers as the final prediction on instances where KG2Code-QA successfully generates executable code, evaluated on the WWQ and LC-QuAD 2.0 datasets. Win indicates that the hard answer achieves a higher F1 score than the soft answer, Lose indicates the opposite, and Tie denotes equal performance.}
  \label{fig winlosetie}
\end{figure}

Moreover, the effectiveness of hard answers generated by KG2Code-QA is investigated by comparing hard and soft answers on instances for which executable programs can be generated. As shown in Figure \ref{fig winlosetie}, hard answers consistently achieve substantially better performance than soft answers, demonstrating their superior reliability and faithfulness for KGQA.

\subsection{Cost Analysis}
This subsection analyzes the time and token costs of different methods. Specifically, using Llama-3.1-8B-Instruct as the backbone model, \textsf{SPARQL-ICL}, \textsf{SPARQL-SFT}, \textsf{Direct Prompt}, \textsf{CoT-Tuning}, and \textsf{KG2Code-QA} are evaluated on WWQ in terms of average prediction time (in seconds), average input token count, and average output token count per question. The detailed results are reported in Table \ref{Table time}. In addition, for GPT-5-Mini-based \textsf{SPARQL-ICL}, \textsf{ToG}, \textsf{GoG}, \textsf{DoG}, \textsf{Direct Prompt}, and \textsf{KG2Code-QA}, the average number of API calls, as well as the average input and output token counts, are further evaluated. Since \textsf{DoG} relies on AgentVerse\footnote{https://github.com/OpenBMB/AgentVerse}, its associated cost is difficult to quantify precisely. Therefore, only the cost excluding AgentVerse is reported. The corresponding results are shown in Table \ref{Table call}.

As shown in Table \ref{Table time}, \textsf{KG2Code-QA} requires less inference time than the training-free baselines \textsf{Direct Prompt} and \textsf{SPARQL-ICL}, although it is slower than \textsf{CoT-Tuning} and \textsf{SPARQL-SFT}. This difference can be attributed to the effect of training, which enables the model to generate target outputs more efficiently and accurately, thereby reducing inference time. However, compared with \textsf{CoT-Tuning}, \textsf{KG2Code-QA} additionally generates code, leading to higher time consumption. \textsf{SPARQL-SFT} incurs the lowest cost, mainly because it only needs to generate short SPARQL queries; however, its overall performance remains unsatisfactory. In terms of token usage, Llama-3.1-8B-Instruct-based \textsf{KG2Code-QA} has the longest average input length, indicating that KG2Code rewriting increases the model input length.

As shown in Table \ref{Table call}, GPT-5-Mini-based \textsf{KG2Code-QA} yields substantially shorter input and output sequences than agent-based methods while requiring significantly fewer API calls. These results suggest that the proposed approach not only achieves superior performance but also demonstrates clear advantages in efficiency.

\begin{table}[h]
\centering
\small
\scalebox{0.9}{
\begin{tabular}{lccc}
    \toprule
     \textbf{Methods} &\textbf{Time} & \textbf{Input Token} & \textbf{Output Token} \\
    \midrule
    \textbf{SPARQL-ICL} &0.88 &291.19 &58.26 \\
    \textbf{SPARQL-SFT} & 0.05 &31.16 &48.79\\
    \textbf{Direct Prompt} & 0.28&470.04 &250.21\\
    \textbf{CoT-Tuning} &0.20 &561.94 &90.04\\
    \textbf{KG2Code-QA (Ours)}  &0.24 &733.02 &156.59\\
    \bottomrule
\end{tabular}
}
\caption{The average time (unit: second), average input token, and average output token per question.}
\label{Table time}
\end{table}

\begin{table}[h]
\centering
\small
\scalebox{0.9}{
\begin{tabular}{lccc}
    \toprule
     \textbf{Methods} &\textbf{Call} & \textbf{Input Token} & \textbf{Output Token} \\
    \midrule
    \textbf{SPARQL-ICL} &1.00 & 292.78&722.82\\
    \textbf{ToG} &9.69 &6141.48 &6972.37\\
    \textbf{GoG} &5.60 & 5948.18& 4491.14\\
    \textbf{DoG-AgentVerse} &2.60 & 2136.08&964.78\\
    \textbf{Direct Prompt}&1.00 &435.53&704.01\\
    \textbf{KG2Code-QA (Ours)} & 1.00&2015.23 &203.88\\
    \bottomrule
\end{tabular}
}
\caption{The average API calls, average input tokens, and average output tokens for baselines based on closed-source LLMs.}
\label{Table call}
\end{table}

\subsection{Scalability Experiments}

\begin{figure}[h]
\centering
  \includegraphics[width=\linewidth]{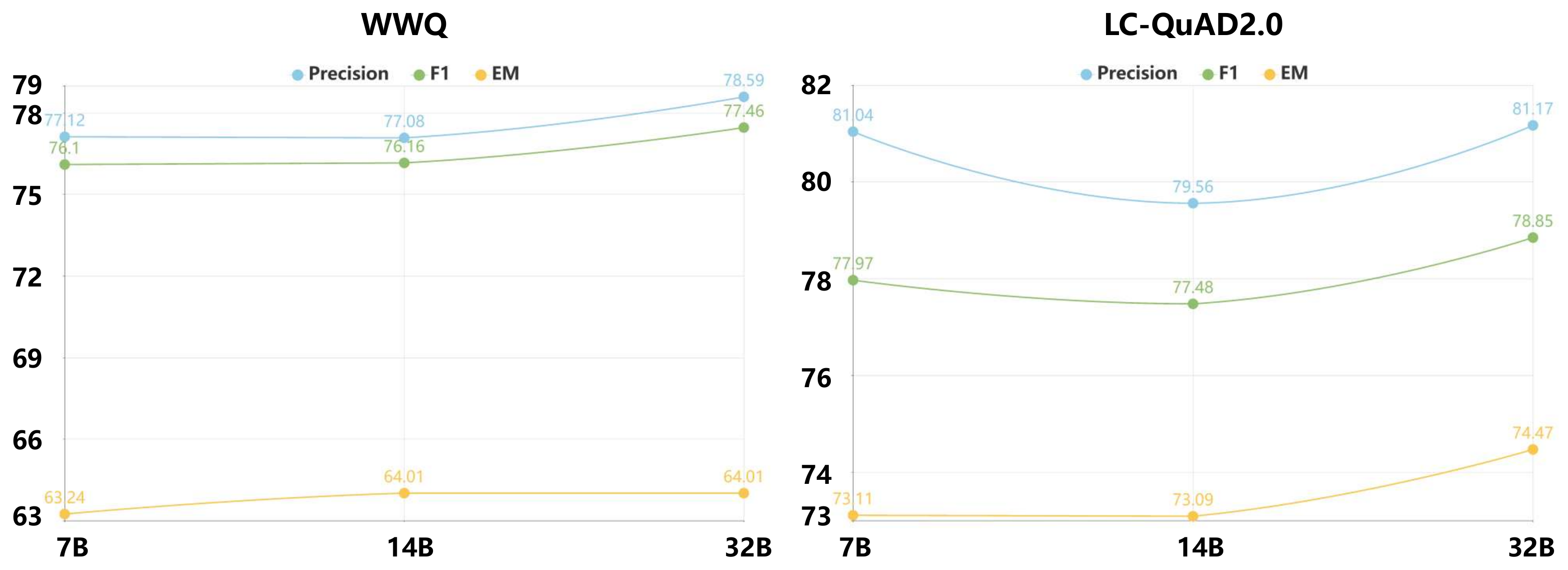}
  \caption{The results of scalability experiments on WikiWebQuestions and LC-QuAD 2.0.}
  \label{fig scale}
\end{figure}

To investigate the impact of model scale on experimental performance, additional experiments are conducted using the Qwen-2.5-Coder series. Specifically, the 14B and 32B variants, which were trained on the same corpus, are evaluated on WikiWebQuestions and LC-QuAD 2.0. The results are presented in Figure~\ref{fig scale}.

The experimental results reveal two main findings. \textbf{(1) KGQA performance exhibits an overall upward trend as model size increases.} Across all evaluated settings, the 32B models achieve the best overall performance, suggesting that larger Qwen2.5-Coder models possess stronger reasoning and knowledge integration capabilities for KGQA. This finding is consistent with the conclusions of prior studies \cite{DBLP:journals/corr/abs-2001-08361,DBLP:conf/esws/HeimMSFD25}. \textbf{(2) The performance gain from 7B to 14B is smaller than that from 14B to 32B, indicating that KGQA places relatively high demands on model capacity.} In particular, the improvement from 7B to 14B is limited, and performance even declines in some cases. By contrast, scaling from 14B to 32B leads to substantially larger and more consistent improvements across the reported metrics. This pattern suggests that mid-sized models may still be insufficient to fully capture the structural reasoning and knowledge grounding required for KGQA, whereas larger models are better able to unlock stronger task performance.

\subsection{Ablation Experiments}

Ablation studies are conducted on LC-QuAD 2.0 to assess the contribution of each component in the proposed framework. Specifically, three variants are evaluated: \textsf{w/o KG2Code}, where the code-based representation is replaced with a textual linearization of triples; \textsf{w/o Code Generation}, where the generated code is removed and only the comments are retained; and
\textsf{w/o Comments}, where the comments are removed and only the code is retained. To ensure a fair comparison, the training data for all ablation settings are constructed from \textsf{KG2Code-QA}, with the same training and validation split adopted across all variants. Because the input or output format differs across these variants, either hard answers or soft answers can be extracted as predictions. Specifically, soft answers are used as the prediction results for \textsf{w/o KG2Code} and \textsf{w/o Code Generation}, whereas hard answers are used for \textsf{w/o Comments}.

The results are reported in Table~\ref{Table ablation}. Removing any component results in a consistent performance drop, highlighting the importance of each module. Notably, removing comments leads to the most severe degradation. This result confirms the effectiveness of the ReAct-style interleaving of reasoning and execution, and further suggests that natural-language comments play a critical role in guiding LLMs to generate coherent and correct code.

\begin{table}[h]
\centering
\small
\scalebox{1}{
\begin{tabular}{lccc}
    \toprule
     \textbf{Methods} &\textbf{Prec} & \textbf{F1} & \textbf{EM} \\
    \midrule
    KG2Code-QA &\textbf{81.58} &\textbf{79.26}&\textbf{74.78}\\
    w/o KG2Code & 79.90&76.63&70.92\\
    w/o Code Generation & 75.61&73.21&68.13\\
    w/o Comments & 65.78&64.41&61.60\\
    \bottomrule
\end{tabular}
}
\caption{The ablation study results on LC-QuAD 2.0.}
\label{Table ablation}
\end{table}

\subsection{Case Study}

This section demonstrates the advantages of KG2Code-QA through a concrete example. \textsf{SPARQL-SFT}, \textsf{Direct Prompt}, and \textsf{GoG} are selected as representative methods of SPARQL-based methods, RAG-based methods, and agent-based methods, respectively, for comparison with KG2Code-QA. 
The specific example is shown in Table \ref{Table case}. \textsf{SPARQL-SFT} generates a SPARQL query whose semantics are inconsistent with the question, instead retrieving individuals whose field of work is pulsar, which leads to incorrect answers. This demonstrates that SPARQL-based approaches remain relatively challenging for LLMs and have substantial limitations. 
\textsf{Direct Prompt} misses one correct answer while introducing an incorrect one, suggesting that RAG-based methods struggle to accurately interpret flattened triples and are prone to hallucination and unfaithfulness. 
Although \textsf{GoG} is able to enumerate triples relevant to the question, it fails to synthesize a complete answer. This result indicates that a substantial gap remains between structured knowledge and its effective utilization by LLMs, and that unfaithfulness continues to be a challenge even when closed-source LLMs are employed. In contrast, KG2Code-QA generates executable code to derive correct and complete answers from the subgraph, thereby demonstrating the effectiveness of code-style representations and showing that KG2Code-QA can effectively mitigate hallucination and unfaithfulness in LLM-based question answering.

\begin{table}[!h]
\centering
\small
\resizebox{\textwidth}{!}{
\begin{tabular}{lp{25cm}}
    \toprule
    \multicolumn{2}{l}{\textbf{Question}: The individual who discovered \textcolor{darkgreen}{pulsar} is in which field of study?}\\
    \multicolumn{2}{l}{\textbf{Answer}: \textcolor{orange}{pulsar}, \textcolor{orange}{astronomy}, \textcolor{orange}{physics}}\\
    \multicolumn{2}{l}{\textbf{Subgraph}: (\textcolor{darkgreen}{pulsar}, discoverer or inventor, Antony Hewish) (Antony Hewish, field of work, \textcolor{orange}{astronomy}) (Antony Hewish, field of work, \textcolor{orange}{physics})}\\
    \multicolumn{2}{l}{\phantom{xxxxxxxxxxx}(\textcolor{darkgreen}{pulsar}, discoverer or inventor, Jocelyn Bell Burnell) (Jocelyn Bell Burnell, field of work, \textcolor{orange}{astronomy}) (Jocelyn Bell Burnell, field of work, \textcolor{orange}{pulsar}) ...}\\
    \midrule
    \textbf{Methods}& \multicolumn{1}{c}{\textbf{Responses}}\\
    \midrule
     \multirow{3}{*}{\textbf{SPARQL-SFT}} & \textbf{[Generated SPARQL]} select distinct ?sbj where \{ ?sbj [field of work] [pulsar] . ?sbj [instance of] [human] \}\newline 
     \textbf{[Executable SPARQL]} select distinct ?sbj where \{ ?sbj wdt:P101 wd:Q4360 . ?sbj wdt:P31 wd:Q5 \}\newline
     \textbf{[Answer]} Jocelyn Bell Burnell | Victoria Kaspi | Arnold Rots | Christian Gouiffés\\
     \cdashline{1-2}
    \multirow{7}{*}{\textbf{Direct Prompt}}& To determine the field of study for the individual who discovered \textcolor{darkgreen}{pulsars}, I will follow these steps:\newline 1. Identify the individuals associated with discovering \textcolor{darkgreen}{pulsars}.\newline 2. Determine their fields of work.\newline From the given subgraph: - Both Antony Hewish and Jocelyn Bell Burnell are mentioned as the discoverers of \textcolor{darkgreen}{pulsars}. - Their fields of work include ``\textcolor{orange}{astronomy}'' and ``\textcolor{orange}{physics}''. Since both individuals are involved in the discovery of \textcolor{darkgreen}{pulsars}, we can conclude that the field of study related to the discovery of \textcolor{darkgreen}{pulsars} includes both \textcolor{orange}{astronomy} and \textcolor{orange}{physics} ... \newline Answer: Astrophysics|\textcolor{orange}{Astronomy}|\textcolor{orange}{Physics}\\
    \cdashline{1-2}
    \multirow{10}{*}{\textbf{GoG}} & \textbf{[Thought]} I need to search \textcolor{darkgreen}{pulsar} to find who discovered it.\newline \textbf{[Action]} Search[\textcolor{darkgreen}{pulsar}]\newline \textbf{[Observation]} Jocelyn Bell Burnell, field of work, \textcolor{orange}{pulsar}|\textcolor{darkgreen}{pulsar}, discoverer or inventor, Antony Hewish|\textcolor{darkgreen}{pulsar}, discoverer or inventor, Jocelyn Bell Burnell\newline \textbf{[Thought]} I need to find the field of study of the discoverer(s) Antony Hewish and Jocelyn Bell Burnell.\newline \textbf{[Action]} Search[Antony Hewish|Jocelyn Bell Burnell]\newline \textbf{[Observation]} Antony Hewish, field of work, \textcolor{orange}{astronomy}|Antony Hewish, field of work, \textcolor{orange}{physics}|Jocelyn Bell Burnell, field of work, \textcolor{orange}{astronomy}|Jocelyn Bell Burnell, field of work, \textcolor{orange}{pulsar} ...\newline \textbf{[Thought]} Antony Hewish and Jocelyn Bell Burnell are the discoverers and their fields of work are astronomy and physics.\newline\textbf{[Action]} Finish[\textcolor{orange}{astronomy}|\textcolor{orange}{physics}]\\
    \cdashline{1-2}
    \multirow{18}{*}{\textbf{KG2Code-QA}}& \# Extract the head entity based on the question from the graph.\newline \# head entity: \textcolor{darkgreen}{pulsar}\newline item = ``\textcolor{darkgreen}{pulsar}''\newline \# Step 1: find the discoverer(s) of \textcolor{darkgreen}{pulsar}\newline discoverers = []\newline for u, v, data in graph.edges(data=True):\newline \phantom{xxxx}if u == item and data.get(``relation'') == ``discoverer or inventor'':\newline \phantom{xxxxxxxx}discoverers.append(v)\newline \# discoverers: [``Jocelyn Bell Burnell'', ``Antony Hewish'']\newline \# Step 2: find the field(s) of work of these discoverers\newline fields = []\newline for u, v, data in graph.edges(data=True):\newline \phantom{xxxx}for u, v, data in graph.edges(data=True):\newline \phantom{xxxxxxxx}if u == d and data.get(``relation'') == ``field of work'':\newline \phantom{xxxxxxxxxxxx}fields.append(v)\newline \# fields: [``\textcolor{orange}{astronomy}'', ``\textcolor{orange}{pulsar}'', ``\textcolor{orange}{physics}'', ``\textcolor{orange}{astronomy}'']\newline return fields\\    
    \bottomrule
\end{tabular}
}
\caption{An example of the responses of different methods.}
\label{Table case}
\end{table}

\subsection{Error Analysis}

To further investigate the sources of error in the proposed framework, 50 erroneous cases are randomly selected and analyzed according to their underlying causes. The errors can be categorized into the following types: \textbf{(1) Subgraph retrieval issues (40\%)}. This category covers cases where the retrieved subgraph fails to cover all correct answers or lacks the information necessary for reasoning toward the answer. \textbf{(2) Dataset issues (18\%)}. These errors arise from changes in the number of ground-truth answers due to updates in Wikidata, ambiguities caused by unclear question semantics, and cases where the questions and answers are entirely unrelated. \textbf{(3) Relation selection errors (12\%)}. This is the primary method-specific error type, where incorrect relations are selected by the LLM during the reasoning process, ultimately leading to wrong results. \textbf{(4) Reasoning errors (10\%)}. Such errors occur when the LLM fails to correctly understand the semantics of the question, or when the relation path corresponding to the question is not explicit and requires additional reasoning. \textbf{(5) Evaluation issues (8\%)}. These cases involve semantically broad questions for which multiple answers are acceptable, while the dataset provides only one reference answer. \textbf{(6) Entity linking errors (4\%)}. In these cases, the head entity in the question differs from the entity in the retrieved subgraph, requiring the LLM to perform entity linking based on its own commonsense knowledge. This suggests that the factual knowledge of the LLM remains imperfect. \textbf{(7) Non-executable code (4\%)}. In this category, the code generated by the LLM is semantically consistent with the question but cannot be executed, causing the system to rely solely on an incorrect soft answer. This indicates that LLMs still exhibit insufficient code generation capabilities and that the proposed method remains affected by hallucination issues. \textbf{(8) Lack of commonsense knowledge (4\%)}. These errors occur when the LLM is required to filter candidate answers according to the semantics of the question. For example, if the question asks for a city, the answer entity should be a city rather than a region. This observation suggests that some questions cannot be answered solely through subgraph operations and instead require additional commonsense knowledge.

\section{Conclusion}

This paper proposes KG2Code, a novel knowledge rewriting strategy that transforms knowledge graphs into executable code. By preserving the structural information of knowledge graphs while naturally aligning with the code-oriented pretraining of LLMs, KG2Code effectively mitigates the structural information loss caused by existing approaches that linearize knowledge graphs into plain text. Building on this code-based representation, a KGQA framework, KG2Code-QA, is further introduced. Specifically, KGQA is formulated as a code generation task, where the model performs step-by-step operations over the retrieved subgraph by generating executable code. This design grounds the reasoning process in the underlying knowledge graph, thereby improving the faithfulness of the generated answers. Compared with SPARQL-based approaches, code imposes fewer constraints and does not require KG-specific training. In addition, an automated pipeline is developed to construct large-scale, high-quality training corpora for training open-source LLMs. The resulting trained open-source models can perform KGQA in a zero-shot setting and demonstrate strong generalization ability. For closed-source LLMs, the proposed method supports training-free inference with only a few demonstrations. Experimental results show that KG2Code-QA significantly outperforms existing KG-enhanced LLM methods for KGQA and generalizes effectively to previously unseen knowledge graphs. Notably, substantial gains are also achieved on closed-source LLMs, further confirming the broad applicability and robustness of the proposed approach.

This work also opens up several promising directions for future research. First, KG2Code-QA can be extended into a graph-based retrieval-augmented generation, or Graph-RAG, framework that supports a wider range of question answering tasks beyond KGQA. In such a framework, a knowledge graph can be constructed from retrieved data sources, such as text and tables, transformed into code, and then used to answer questions through the proposed method. Second, code-based representations are not limited to question answering and may also benefit other complex reasoning tasks, such as knowledge graph completion and knowledge editing. Finally, the framework can be extended to multimodal settings, where different interfaces can be implemented in code to facilitate the integration and reasoning of multimodal information.


\appendix

\section{Question Types}
\label{appendix type}
Six common types of KGQA are summarized, with detailed descriptions presented in Table \ref{Table type}.

\begin{table}[!htbp]
\centering
\small
\label{tab:question_types}
\begin{tabular}{p{3cm} p{5cm} p{3.5cm}}
\toprule
\multicolumn{1}{c}{\textbf{Question Type}} & \multicolumn{1}{c}{\textbf{Description}} & \multicolumn{1}{c}{\textbf{Example}} \\
\midrule

Length Calculation 
& Questions that require counting the number of items in the final query result. 
& How many kids were there in the kennedy family?\\
\midrule
Boolean Evaluation 
& Questions whose answers require applying logical operations such as AND or OR
&  Is it true that in Dubai there are 540 cars and trucks per thousand of population?\\
\midrule
Numerical Comparison 
& Questions that require filtering candidate answers according to numerical attributes and comparison conditions. 
& Which cities in China have a population of more than ten million?\\
\midrule
List Sorting 
& Questions that ask for an extremum or rank-based answer according to a sortable attribute or score. 
&What Han Chinese person has the largest net worth?\\
\midrule
Set Operations 
& Questions that require combining or intersecting multiple query results. 
& what and where did Donald Trump consider? \\
\midrule
String-level Manipulation 
& Questions that depend on string-based constraints or textual patterns in the candidate answers. 
& What stellar classification starts with the letter ``t''?\\ 
\bottomrule
\label{Table type}
\end{tabular}
\caption{The description of six question types.}
\end{table}

\section{Examples for KG2Code-QA}
\label{appendix examples}
The code input and code output of KG2Code-QA are presented in this section. For factual, counting, and boolean questions, one representative example from each category is provided in Table~\ref{Table fact}, Table~\ref{Table count}, and Table~\ref{Table bool}, respectively.

\begin{table}[htbp]
\small
\scalebox{1}{
\begin{tcolorbox}
    \textbf{KG2Code-QA for Factual Question}\\
    \textcolor{orange}{\textbf{[Instruction]}}\\
    \includegraphics[width=\linewidth,keepaspectratio]{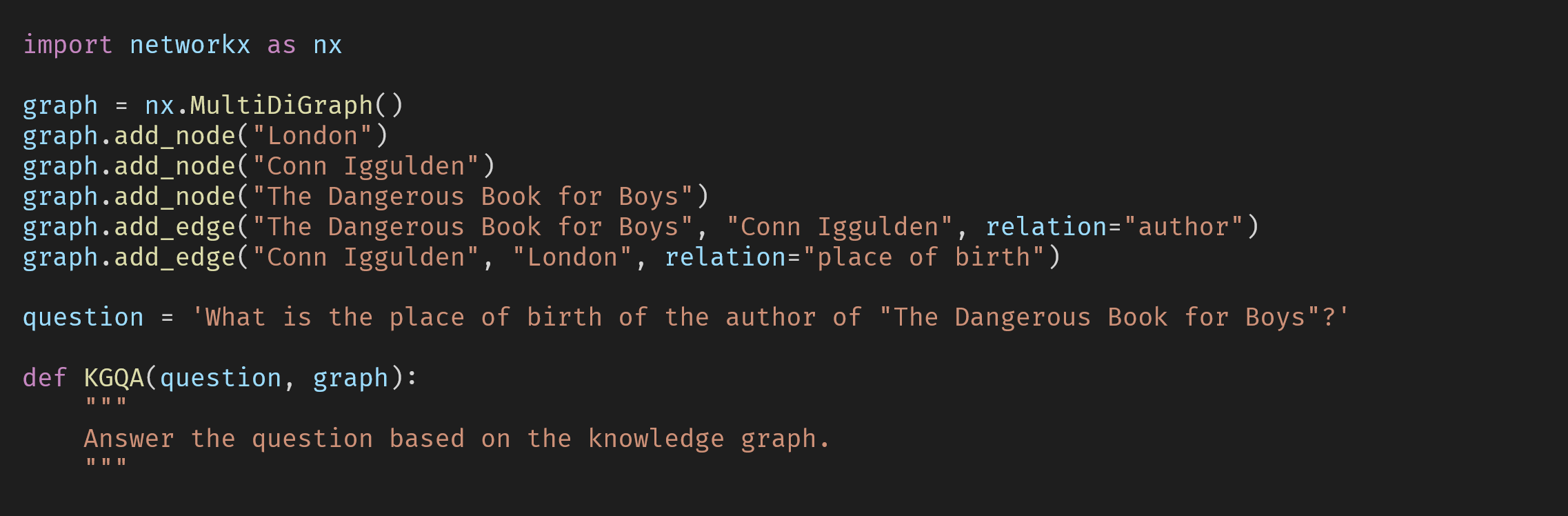}
    \textcolor{darkgreen}{\textbf{[Output]}}\\
    \includegraphics[width=\linewidth,keepaspectratio]{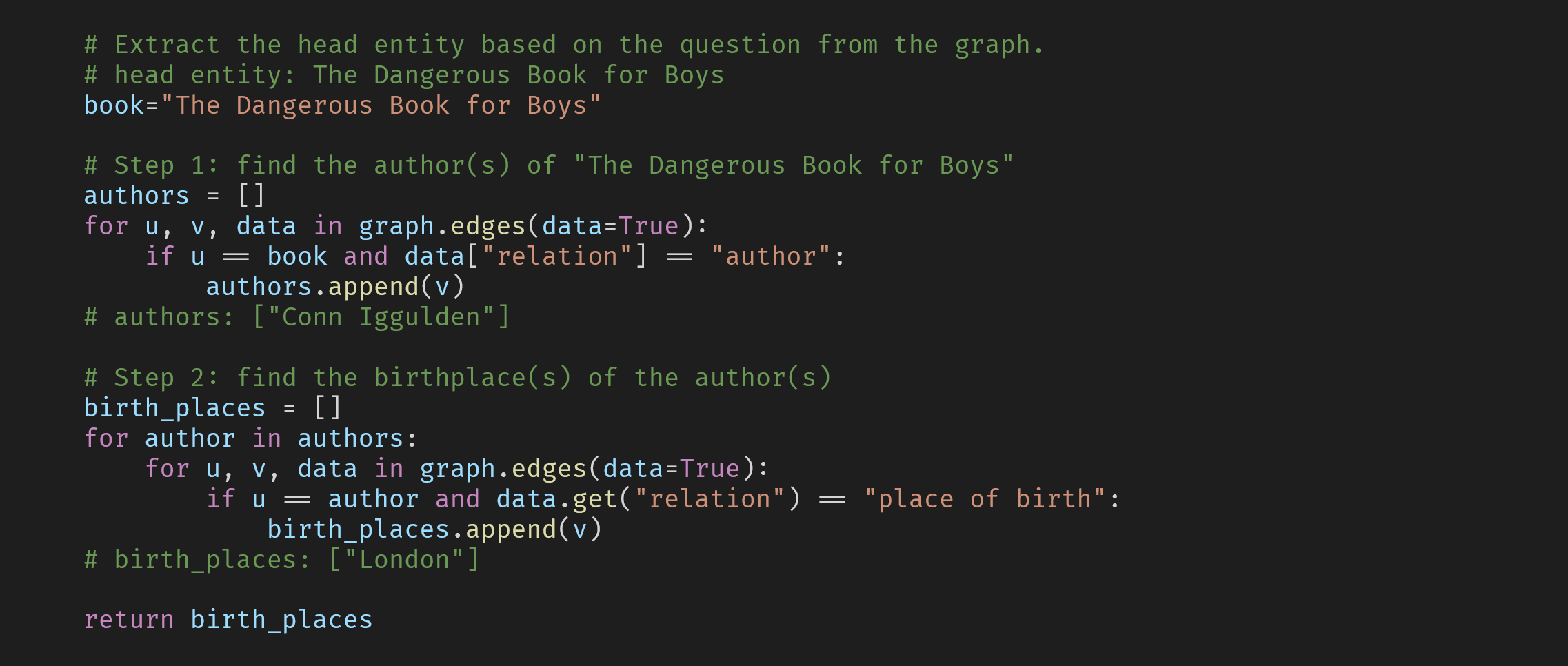}
\end{tcolorbox}
}
\caption{One example of factual question.}
\label{Table fact}
\end{table}

\begin{table}[htbp]
\small
\scalebox{1}{
\begin{tcolorbox}
    \textbf{KG2Code-QA for Counting Question}\\
    \textcolor{orange}{\textbf{[Instruction]}}\\
    \includegraphics[width=\linewidth,keepaspectratio]{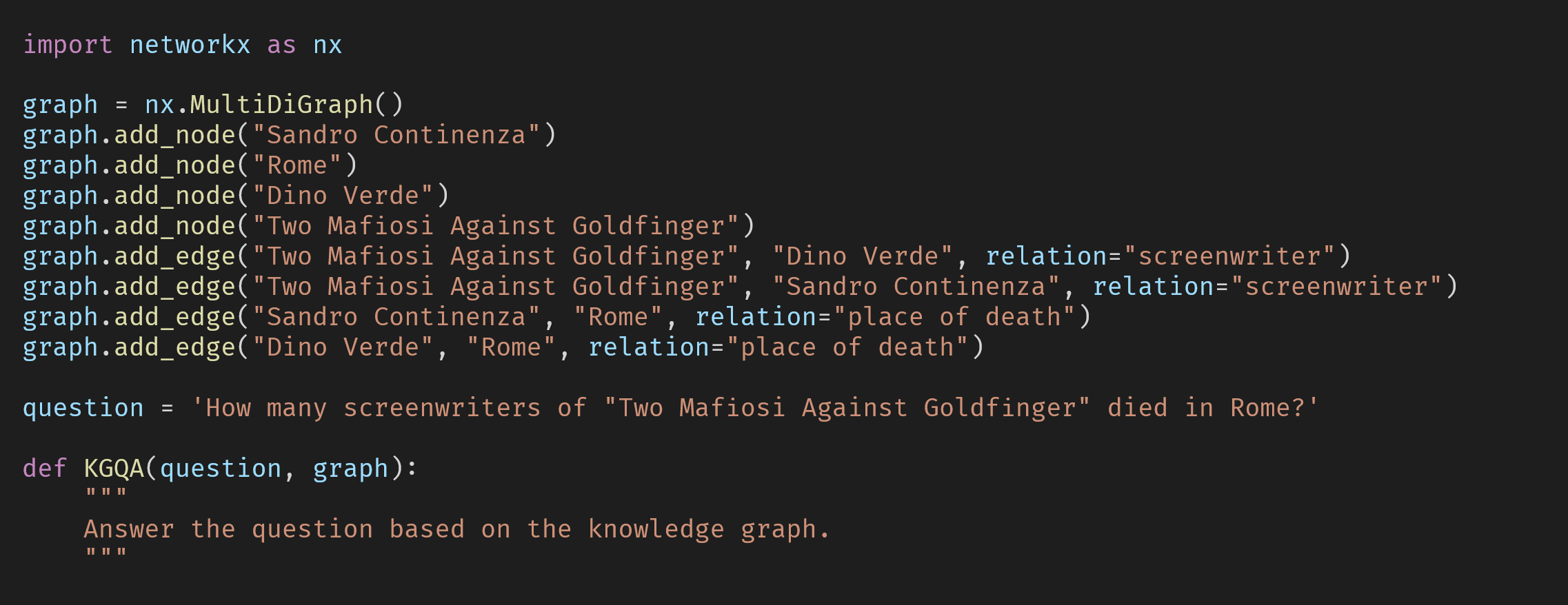}
    \textcolor{darkgreen}{\textbf{[Output]}}\\
    \includegraphics[width=\linewidth,keepaspectratio]{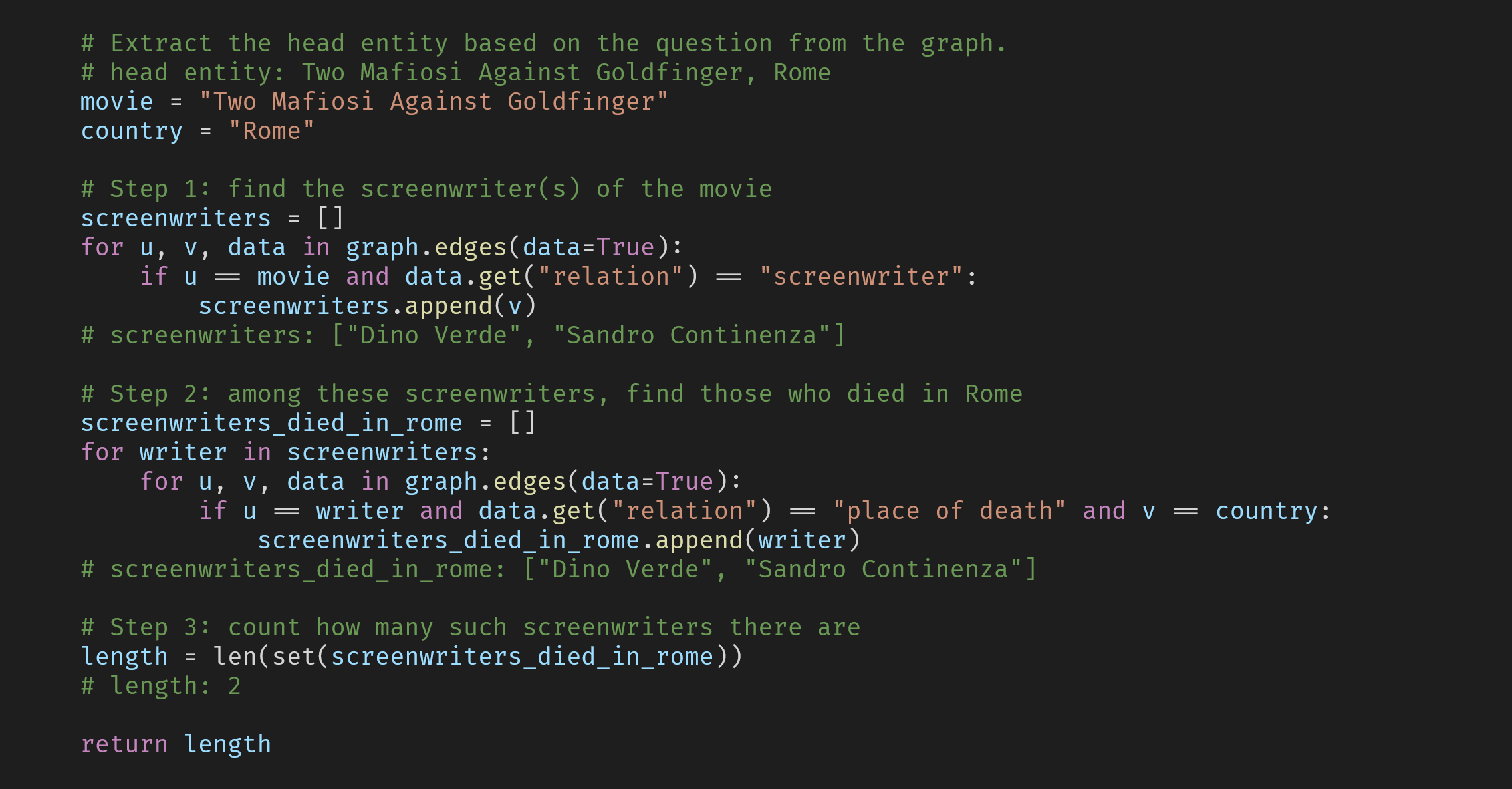}
\end{tcolorbox}
}
\caption{One example of counting question.}
\label{Table count}
\end{table}

\begin{table}[htbp]
\small
\scalebox{1}{
\begin{tcolorbox}
    \textbf{KG2Code-QA for Boolean Question}\\
    \textcolor{orange}{\textbf{[Instruction]}}\\
    \includegraphics[width=\linewidth,keepaspectratio]{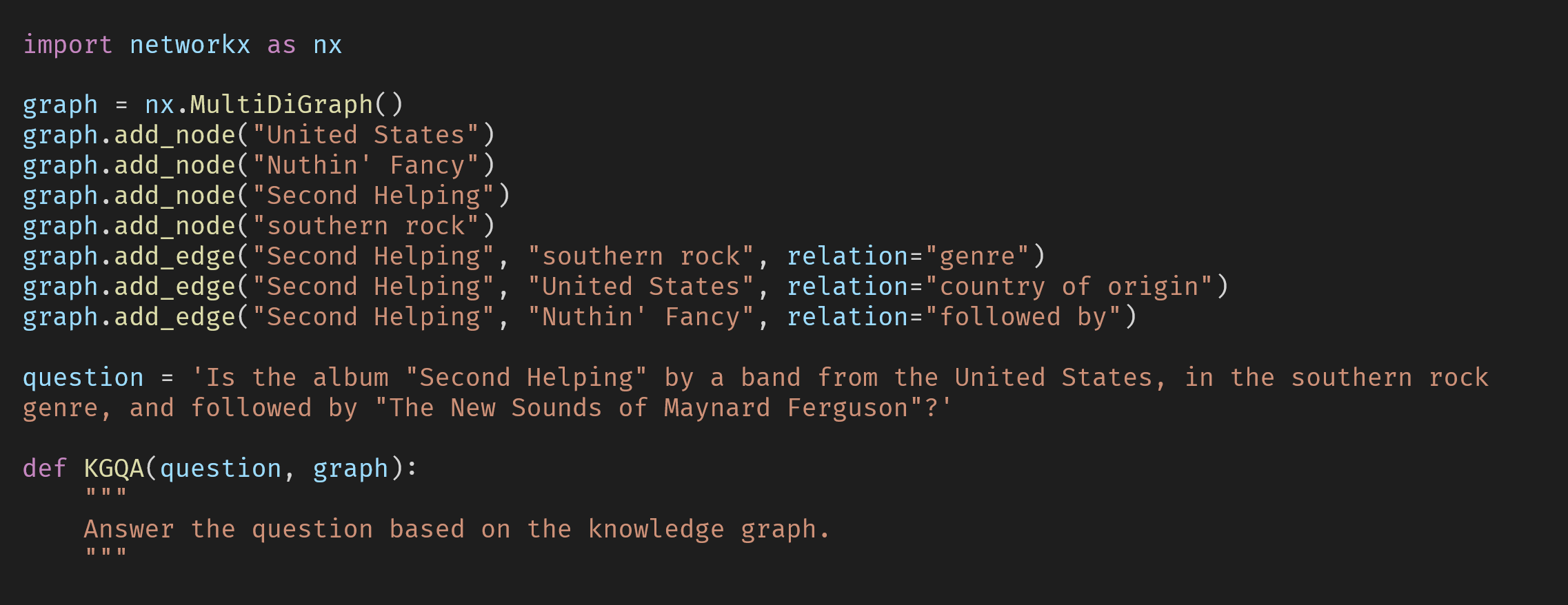}
    \textcolor{darkgreen}{\textbf{[Output]}}\\
    \includegraphics[width=\linewidth,keepaspectratio]{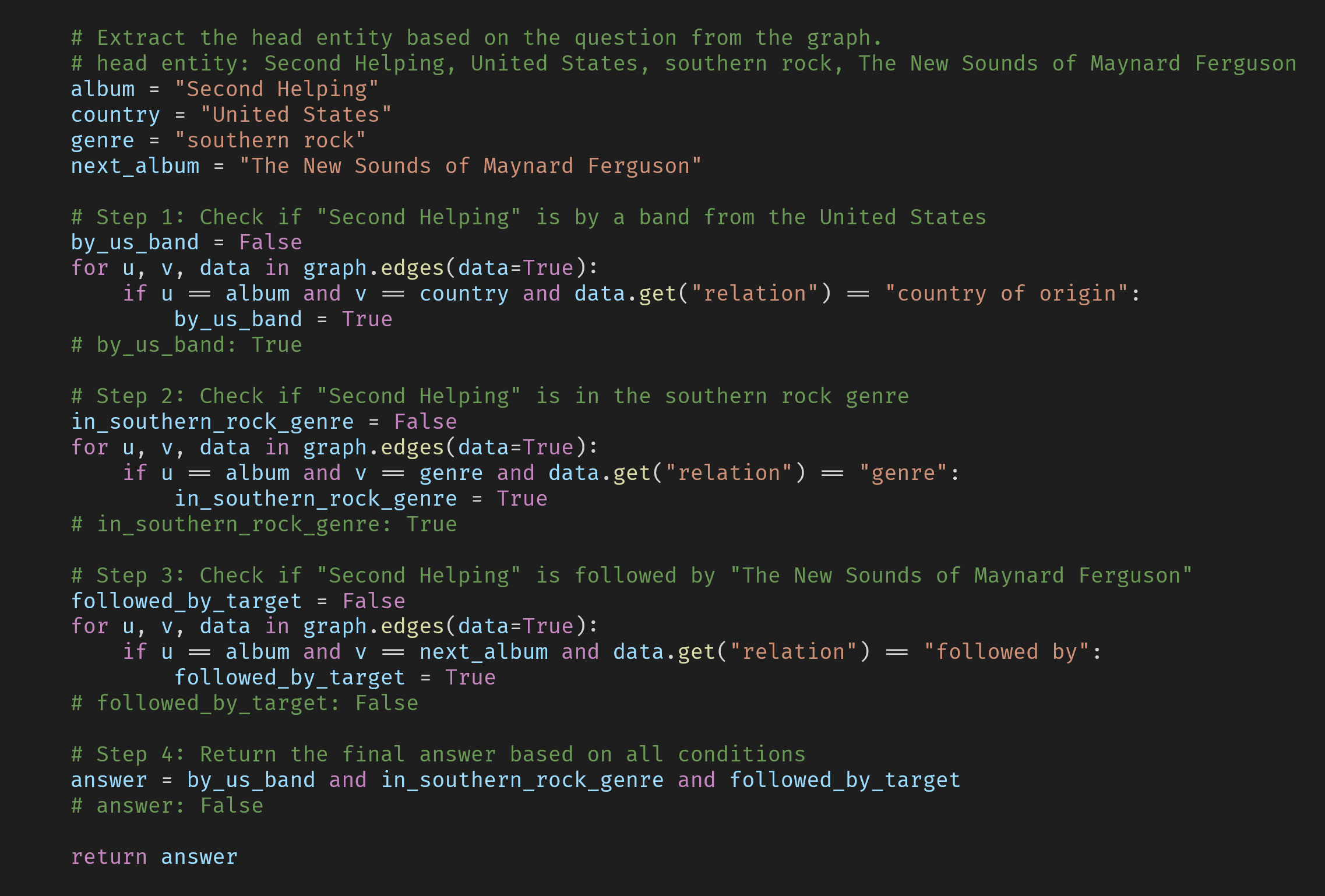}
\end{tcolorbox}
}
\caption{One example of boolean question.}
\label{Table bool}
\end{table} 

\section{Prompt Templates}
\label{appendix prompt templates}
This section presents the prompt templates used in this work. The templates corresponding to different methods are introduced first. For the SPARQL-based method, the question-to-SPARQL prompt template is shown in Table \ref{Table question2sparql}. ToG\footnote{https://github.com/DataArcTech/ToG}, GoG\footnote{https://github.com/YaooXu/GoG}, DoG\footnote{https://github.com/mira-ai-lab/DoG}, and G-Retriever\footnote{https://github.com/XiaoxinHe/G-Retriever} are implemented based on their original source code on GitHub. For the \textsf{Direct Prompt} and \textsf{CoT-Tuning} baselines, the question-answering prompt template is provided in Table \ref{Table question-answering}. For \textsf{KG-to-Text}, \textsf{Summary}, and \textsf{CoTKR}, the same knowledge rewriting prompt templates as those in \cite{DBLP:conf/emnlp/WuHHHQ0P24} are adopted. Their question-answering template is shown in Table \ref{Table question-answering}. In addition, the prompt used to generate questions from SPARQL queries during corpus generation is presented in Table~\ref{Table generate question}.

\begin{table}[htbp]
\small
\scalebox{0.9}{
\begin{tcolorbox}
    \textbf{Question-to-SPARQL Prompt in SPARQL-based Methods}\\
    \textcolor{orange}{\textbf{[Instruction]}}\\
    Please generate the corresponding SPARQL query based on the given question.\\
    \textbf{Question: \{question\}}\\
    \textbf{Sparql:}\\
\end{tcolorbox}
}
\caption{Question-to-SPARQL Prompt in SPARQL-based Methods.}
\label{Table question2sparql}
\end{table}

\begin{table}[htbp]
\small
\scalebox{0.9}{
\begin{tcolorbox}
    \textbf{Question-Answering Prompt in Direct Prompt/CoT-Tuning Baselines}\\
    \textcolor{orange}{\textbf{[Instruction]}}\\
    Please answer the question based on the subgraph retrieved from the knowledge graph. First, provide your Chain-of-Thought (CoT) reasoning process. At the end, list all answers in a single line, starting with ``Answer: '' and separating each answer by ``\textbar'' as follows: Answer: First answer\textbar Second answer\textbar Third answer ...\\
    \textbf{Subgraph: \{subgraph\}}\\
    \textbf{Question: \{question\}}\\
    \rule{\linewidth}{0.2mm}\\
    \textbf{Question-Answering Prompt in KG-to-Text/Summary/CoTKR Baselines}\\
    \textcolor{orange}{\textbf{[Instruction]}}\\
    Please answer the question based on the following knowledge. First, provide your Chain-of-Thought (CoT) reasoning process. At the end, list all answers in a single line, starting with ``Answer: '' and separating each answer by ``\textbar" as follows: Answer: First answer\textbar Second answer\textbar Third answer ...\\
    \textbf{Knowledge: \{knowledge\}}\\
    \textbf{Question: \{question\}}
\end{tcolorbox}
}
\caption{Question-answering prompt in Direct Prompt/CoT-Tuning/KG-to-Text/Summary/CoTKR Baselines.}
\label{Table question-answering}
\end{table}

\begin{table}[htbp]
\small
\scalebox{0.9}{
\begin{tcolorbox}
    \textbf{Question Generation Prompt}\\
    \textcolor{orange}{\textbf{[Instruction]}}\\
    Please generate a natural language question that corresponds to the SPARQL query below. In the query, CVT nodes represent blank nodes used to model n-ary relations. When generating the question, you should reflect the semantic role of these blank nodes, but you must not mention CVT or blank nodes explicitly in the question.\\
    \textbf{\{sparql\}}\\
    \textbf{Question:}\\
\end{tcolorbox}
}
\caption{Question Generation Prompt for corpus construction.}
\label{Table generate question}
\end{table}

\section{Acknowledgments}
This work is partially supported by National Nature Science Foundation of China under No. 62476058. We thank the Big Data Computing Center of Southeast University for providing the facility support on the numerical calculations in this paper.

\section{Data availability}

Our code and data are available at Github: https://anonymous.4open.science/r/test-5074519-5563074/.

\section{Declaration of generative AI and AI-assisted technologies in the manuscript preparation process}

During the preparation of this work, the author(s) used ChatGPT by OpenAI to assist with language polishing, improving readability, organizing content, and drafting/revising portions of the manuscript. After using this tool/service, the authors carefully reviewed, verified, edited, and adapted the content as needed, and take full responsibility for the content of the published article.



\bibliographystyle{elsarticle-num} 
\bibliography{references_no_urls}



\end{document}